\documentclass[conference]{IEEEtran}
\IEEEoverridecommandlockouts
\usepackage{cite}
\usepackage{amsmath,amssymb,amsfonts}
\usepackage{algorithmic}
\usepackage{graphicx}
\usepackage{textcomp}
\usepackage{xcolor}

\usepackage{multirow}

\def\BibTeX{{\rm B\kern-.05em{\sc i\kern-.025em b}\kern-.08em
    T\kern-.1667em\lower.7ex\hbox{E}\kern-.125emX}}
\begin{document}

\title{MorphGuard: Morph Specific Margin Loss for Enhancing Robustness to Face Morphing Attacks}


\author{Iurii Medvedev\\
\textit{$^1$Institute of Systems}\\
\textit{and Robotics,}\\
\textit{University of Coimbra,}\\
Coimbra, Portugal\\
{\tt\small iurii.medvedev@isr.uc.pt}
\and
Nuno Gonçalves $^{1,2}$\\
\textit{$^2$Portuguese Mint and Official}\\
\textit{Printing Office (INCM),}\\
\textit{Lisbon, Portugal}\\
{\tt\small nunogon@deec.uc.pt}
}

\IEEEoverridecommandlockouts \IEEEpubid{\makebox[\columnwidth]{979-8-3503-5447-8/24/\$31.00 ©2024 European Union \hfill} \hspace{\columnsep}\makebox[\columnwidth]{ }}

\maketitle

\IEEEpubidadjcol

\begin{abstract}

Face recognition has evolved significantly with the advancement of deep learning techniques, enabling its widespread adoption in various applications requiring secure authentication. However, this progress has also increased its exposure to presentation attacks, including face morphing, which poses a serious security threat by allowing one identity to impersonate another. Therefore, modern face recognition systems must be robust against such attacks.  

In this work, we propose a novel approach for training deep networks for face recognition with enhanced robustness to face morphing attacks. Our method modifies the classification task by introducing a dual-branch classification strategy that effectively handles the ambiguity in the labeling of face morphs. This adaptation allows the model to incorporate morph images into the training process, improving its ability to distinguish them from bona fide samples.  

Our strategy has been validated on public benchmarks, demonstrating its effectiveness in enhancing robustness against face morphing attacks. Furthermore, our approach is universally applicable and can be integrated into existing face recognition training pipelines to improve classification-based recognition methods.
   
\end{abstract}

\section{Introduction}


Digitalization and technological advancements have led to an increasing demand for automated security solutions in various authentication processes, such as entry-exit systems, mobile device unlocking, and transaction verification. To meet the needs of modern lifestyles, these digital solutions must be both fast and highly secure. In recent years, the rapid evolution of deep learning technologies has significantly impacted modern security measures. Such approaches often rely heavily on various biometric modalities. Among them, facial recognition has gained particular attention and widespread adoption due to its ease of biometric sample acquisition (face image) and recent advancements in computer vision techniques.



A special place in the variety of security solutions is taken by document security applications. These solutions require reliable methods to verify the authenticity of a face image, confirm its association with a specific individual, or identify the legitimate document owner. Security measures can be implemented at multiple stages, including document issuance (during face image enrollment), differential verification against a live reference (such as in automated border control systems), and subsequent document verification in cases requiring live enrollment.  

Face Recognition Systems (FRSs) are widely employed in security applications, identification and verification processes across various domains. However, despite their advancements, FRSs remain vulnerable to attacks, particularly from sophisticated image manipulation techniques designed to deceive the system.


Face morphing is a technique that involves blending two or more facial images to generate a synthetic image that retains biometric characteristics of the original subjects. This manipulation can lead to security vulnerabilities, as the resulting morphed image may match multiple individuals, enabling unauthorized access to identity verification systems. A major concern is the potential exploitation of face morphing for the creation of fraudulent identity documents, which, if undetected, could be used for illicit activities. Although some of these forged documents are occasionally identified during border control inspections, the actual prevalence of such fraud remains uncertain \cite{torkar2023morphing}.  

Despite significant advancements in face recognition driven by deep learning, face morphing remains a persistent threat, challenging both automated recognition systems and human-based verification processes. Addressing this issue is critical for ensuring the reliability and security of biometric identification systems. Developing robust countermeasures against morphing attacks is essential to prevent illegitimate usage of identity documents.


%
In this work, we revisit classification approaches for face recognition in order to increase the robustness to face morphing attacks by incorporating face morph images into the training process. We propose a novel training framework based on sample-specific modifications, introducing distinct marginal penalization for face morph samples. This approach enhances the separation between morphs and bona fide identities, improving recognition performance. 
In general, this work addresses the question of how to integrate and utilize face morphs in the training process of a face recognition network.

In our experiments, we mainly train models from scratch, however, we also demonstrate that our method can also be applied to enhance the robustness of previously trained and deployed models. We evaluated our approach on a public benchmark, and the results demonstrate that our strategy effectively improves the robustness of face recognition networks against morphing attacks.

\section{Related Work}

Our approach in this work builds upon recent advancements in face recognition and face morphing detection. In this section, we review several studies in these fields and discuss their relevance to our proposed method.

\subsection{Face Recognition}

Face recognition technology has seen remarkable advancements due to the rapid progress in deep learning techniques. These advancements have led to more accurate and efficient recognition systems, widely used in biometric security, surveillance, and authentication applications. Central to this progress is the development of discriminative feature representations that enable robust identity verification across varying conditions. This section discusses the impact of deep learning on face recognition, with a focus on metric learning approaches and classification-based strategies.

Deep learning, particularly convolutional neural networks (CNNs), has significantly enhanced face recognition capabilities by learning highly discriminative representations from face images \cite{ImageNet_cite}. CNNs provide an effective means of extracting meaningful facial features from unconstrained images, making them particularly useful in biometric applications. The training of such networks generally involves large-scale datasets, often collected under real-world conditions, which improves the system’s robustness to variations in pose, illumination, and occlusion.  

By optimizing feature representations, deep learning models enable efficient identity discrimination in a low-dimensional feature space. These models commonly rely on similarity metrics such as cosine or Euclidean distance to measure how closely two face images match. While CNN-based architectures have become the standard for face recognition, different training strategies exist, primarily categorized into metric learning and classification-based approaches.

Metric learning methods optimize the feature space by directly enhancing the similarity measure between face embeddings. Unlike classification-based approaches, metric learning does not rely on predefined class labels but instead focuses on maximizing inter-class variance and minimizing intra-class differences \cite{chopra_metric_paper}. One of the pioneering methods in this category is FaceNet, which introduced triplet loss \cite{facenet}. This method optimizes embeddings using a triplet of samples: an anchor image, a positive sample (same identity), and a negative sample (different identity). The goal is to minimize the anchor-positive distance while maximizing the anchor-negative distance.  

Despite their advantages, metric learning approaches require sophisticated sample mining strategies to ensure stable convergence during training. Sampling hard positives and hard negatives is crucial to improving discrimination power, and poorly chosen samples can lead to suboptimal training performance \cite{working_hard}. To address these challenges, modern contrastive learning techniques incorporate adaptive margin constraints and clustering-based strategies to refine the embedding space \cite{coreface}.

The most commonly used strategy for training deep face recognition models is classification-based learning, which treats face recognition as a multi-class classification problem. These methods train networks using identity-labeled datasets and typically employ Softmax loss or its modifications \cite{deepid_paper, deepid2_paper, deepid2_plus_paper}. The learned feature space is then used for open-set recognition tasks, where unseen identities must be distinguished.  

To further enhance recognition performance, various modifications have been introduced to improve intra-class compactness and inter-class separability. Several methods focus on enforcing additional constraints, such as center loss \cite{centerface_paper}, which pushes intra-class features toward their mean center, and margin-based Softmax variations like SphereFace \cite{sphereface_paper}, CosFace \cite{cosface_paper}, and ArcFace \cite{arcface_paper}. These methods introduce angular or additive margin constraints to enhance feature discrimination.  

Recent works have also explored sample-specific strategies to refine feature learning. Hard sample mining techniques \cite{npcface} and uncertainty-based learning \cite{probabilistic_embedding} help adjust the training process dynamically based on sample difficulty. Additionally, quality-aware face recognition methods, such as MagFace \cite{Magface}, QualFace \cite{QualFace, QualFace2}, and AdaFace \cite{AdaFace}, leverage image quality metrics to optimize feature representations. These approaches incorporate quality-aware losses that adapt sample importance based on factors such as resolution, occlusion, and pose variations, ultimately improving recognition robustness.

Summarizing, both metric learning and classification-based approaches offer effective solutions, classification-based methods remain the dominant strategy due to their scalability and stability in large-scale training scenarios. However, metric learning techniques provide valuable insights into improving feature representations and remain essential for specialized applications, such as transfer learning and few-shot recognition. Future research is expected to focus on improving sample efficiency, adapting face recognition models to extreme conditions, and enhancing their generalization to unseen data.

\subsection{Face Morphing}


To create an image for a presentation attack, an impostor typically uses their own face as a base. The target identity’s image is then used to extract key facial features, which are subsequently transferred onto the impostor’s face. Basic blending techniques alone can already produce deceptive results capable of fooling face recognition systems. However, this process has traditionally required extensive manual retouching. With recent advancements, new methods now enable the generation of high-quality, artifact-free morphs through largely automated pipelines.



Early face morphing techniques \cite{magic_passport} \cite{ubo_morpher} primarily relied on facial landmark detection to align images and blend them in the spatial domain. However, these landmark-based approaches often introduced visible blending artifacts due to inaccuracies in landmark detection.  

With advancements in generative deep learning, more sophisticated face morphing techniques have emerged, leveraging deep latent feature representations. Generative Adversarial Networks (GANs) \cite{NIPS2014_5ca3e9b1}, particularly encoder-decoder architectures, have been widely used for this purpose. For instance, MorGAN \cite{morGAN} aims to produce morphs that resemble real images while maintaining diversity among generated samples. StyleGAN-based methods \cite{styleGAN} enable high-quality morphing by interpolating latent facial representations, achieving seamless and realistic results. MIPGAN \cite{MIPGAN_morphing_paper} further refines this approach by optimizing the StyleGAN latent space with a novel loss function to enhance identity preservation in morphed images. Another modification of GAN-based morphing was introduced with the conditioning of the latent domain (codebook learning), which results in a discrete and compact latent representation for enabling a high-quality face morphing generation \cite{morcode}.  MorDIFF \cite{MorDIFF} utilizes diffusion autoencoders to generate smooth, high-fidelity morphs. The most challenging results of this strategy can be obtained with the brute force search \cite{greedydim}. Also Diffusion Model based approaches can be improved by conditioning of biometric template inversion \cite{ladimo}. ReGenMorph \cite{ReGenMorph} integrates image-level morphing with GAN-based synthesis to produce visually convincing morphed faces with minimal artifacts.

A significant amount of recent research has focused on detecting face morphs within biometric security systems. Various Morphing Attack Detection (MAD) techniques have been developed and explored, particularly in both no-reference (S-MAD) and differential (D-MAD) scenarios.  

Early S-MAD methods relied on handcrafted image features such as Binarized Statistical Image Features (BSIF) \cite{7791169}, Local Binary Pattern (LBP) \cite{OJALA199651}, Local Phase Quantization (LPQ) \cite{Blur_Texture_Classification}, and Photo Response Non-Uniformity (PRNU) \cite{Morphing_detection_PRNU} to analyze sensor noise and detect morphing artifacts. More recently, deep learning approaches have gained traction for face morphing detection. For instance, OrthoMAD introduces a regularization technique that enforces orthogonal latent vectors to separate identity information from morphing traces \cite{orthomad}. MorDeephy employs a fused classification strategy to enhance generalized morphing detection \cite{MorDeephy}. MADation approach \cite{madation} introduces a framework that adapts pre-trained foundation models, specifically CLIP \cite{clip}, for single image morphing attack detection. Additionally, methods such as \cite{Tapia2021SingleMA} incorporate few-shot learning using Siamese networks and domain generalization techniques, leveraging a triplet-semi-hard loss function and clustering mechanisms for class assignment.

In contrast, D-MAD techniques utilize trusted reference samples to identify presentation attacks by detecting inconsistencies between the live capture and the document image. For example, Borghi et al. proposed a differential morphing detection framework that fine-tunes pretrained networks within an identity verification and artifact detection pipeline \cite{Double_Siamese_Morphing}. Qin et al. introduced a feature-wise supervision approach for detecting and localizing morphing attacks in both single-image and differential setups, enabling detailed characterization of morphing patterns \cite{FMD_Feature_Wise_Supervision}. Hybrid approaches to Differential Morphing Attack Detection were proposed that integrate identity-based cues from D-MAD with artifact-sensitive features from S-MAD \cite{dmad_combining_identity_features}.  Furthermore, Ferrara et al. explored an alternative differential approach by applying face demorphing techniques, in which a trusted live capture is used to reverse morphing artifacts and reveal the authentic identity of the document holder \cite{face_demorphing}. The demorphing approach can be extended with the deep learning tools. For instance, IFT-Net (Identity Feature Transfer Network). framework \cite{iftnet} relies on symmetric dual-network architectures to reverse-engineer the genuine face.

\subsection{Robustness to Morphing Attacks}

An alternative approach to direct face morphing detection is to enhance the robustness of biometric templates against morphing attacks. This involves implementing protective measures at every stage of the biometric security system, ensuring that the process of extracting biometric templates remains secure from potential vulnerabilities.  

The stability and robustness of facial templates (representations) are critical concerns that have been explored in several studies \cite{Biometric_Systems_under_Morphing_Attacks}. For example, Marriott et al. \cite{Marriott2020RobustnessOF} evaluated multiple facial recognition algorithms for their resistance to face morphing attacks, highlighting the substantial threat posed by such manipulations.  

To address these challenges, various metric-learning-based methods have been introduced to combat morphing attacks. Some of these approaches propose dedicated branches for face morphs within Siamese deep architectures, ensuring controlled feature distribution for morphed faces \cite{medvedev2024quadruplet}.  

To address the numerical evaluation of robustness against morphing attacks several benchmarks were introduced. They usually adopt the Mated Morph Presentation Match Rate (MMPMR) metrics \cite{Biometric_Systems_under_Morphing_Attacks}, which define a successful morphing attack as one in which all contributing identities are matched with the morph. For instance NIST FRTE MORPH Benchmark have started incorporating robustness evaluation metrics \cite{bench_NIST_morph}.  Some studies have utilized the FRGC\_V2 dataset \cite{FRGC_V2_dataset, iwbf1, iwbf2} to measure the vulnerability of face recognition systems to morphing attacks using the MMPMR metric. More recently, new benchmarking frameworks have been proposed to compare different face recognition methods and analyze the deception potential of various morphing techniques \cite{morfacing}.  

Our work approaches the issue of increasing this robustness by adopting classification-based methods to protect the feature domain of face recognition network against morphing attacks.


\begin{figure*}
\begin{center}
  \includegraphics[width=0.85\linewidth]{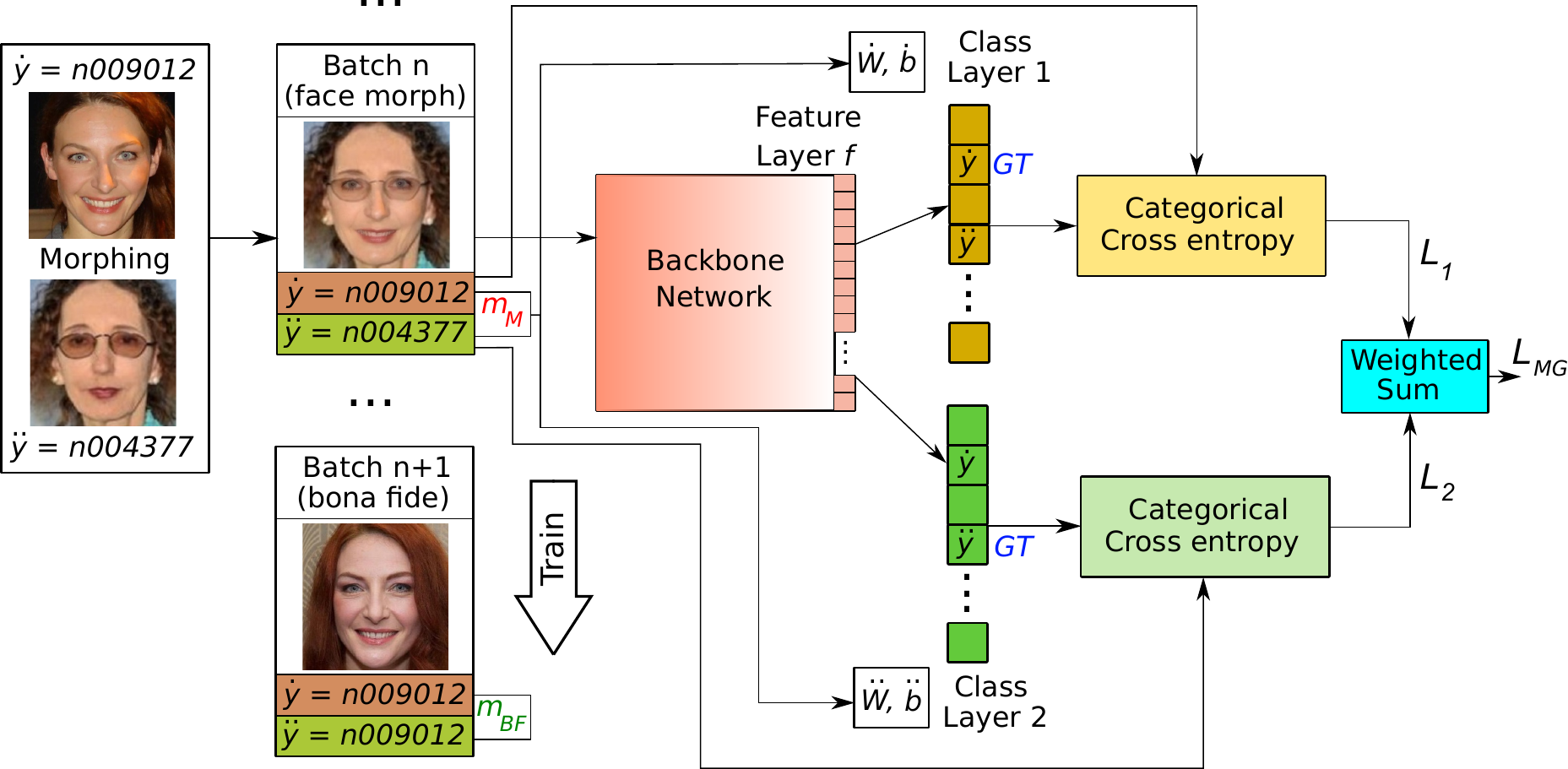}
\end{center}
   \caption{Schematic of the proposed training method. For simplicity of visualization batch contains a single image. In case of Morph sample Loss Function for Class Layers 1 and 2 is computed with different margin for $\dot{y}$ and $\ddot{y}$ classes. In case of Bona Fide Sample the $\dot{y}$ and $\ddot{y}$ are equal and ground truth, thus margins are not applied for their respective classes.  Labels $\dot{y}$ and $\ddot{y}$ are indicated by names (the real setup utilizes their numerical index value, which is encoded to one-hot vector).}
\label{fig:morphguard_schematic}
\end{figure*}

\section{Methodology}




Our strategy for increasing robustness addresses the specific separation of deep features in the latent feature domain. Technically, our goal is to find a learning strategy that incorporates face morphs into the training pipeline and allows a better separation of those morphs from their original subjects.  

We follow the common approach of training a deep network for face recognition through an identity classification based task. Taking into account the nature of the face morphing operation, a straightforward classification definition here is ambiguous, since a morph sample belongs to multiple classes. We seek a universal strategy that can be implemented for a standalone training dataset with the potential to be adopted in common pipelines for training face recognition networks. We define the following target properties of our approach:  

\begin{itemize}
\item{The methodology should follow the classification approach and allow the processing of both morph and bona fide images from the same dataset. }
\item{ Each input image must be classified explicitly and unambiguously. } 
\end{itemize}

Our approach is based on the coupling of classification layers, where such an ambiguity is avoided by defining separate classification branches with distinct labeling.

We start our formulation with the common definition of the softmax loss function, which typically serves as the basis for most recently developed loss functions in face recognition. The softmax loss is generally formulated as follows:

\begin{equation}
    L_{Softmax} = \frac{1}{N}\sum_{i} -\log (\frac{e^{f_{y_i}}}{ \sum_{j}^{C} e^{f_{y_j}}}).
\end{equation}

In this context, $C$ represents the total number of identities (or classes), $N$ denotes the batch size (i.e., the number of samples in a batch), $y_i$ corresponds to the class index of the $i$-th sample, and $f_{y_j}$ refers to the $y_j$-th element of the logits vector $\mathbf{f}$ in the final layer.

L2 normalization is typically applied to both the weight vectors $\mathbf{W_j}$ and the deep features $\mathbf{x_i}$ to constrain them to lie on a hyper-sphere in $\mathbb{R}^d$ space, where $d$ represents the dimensionality of the feature vector $\mathbf{f}$. Under this normalization, $f_{y_j}$ can be expressed as $f_{y_j} = W_j^T x_i = \cos(\theta_j)$, where $\theta_j$ is the angle between $\mathbf{W_j}$ and $\mathbf{x_i}$ and bias $b_j = 0$. Indeed, this normalization naturally suggests to use of an angular similarity metric for distinguishing between different samples.



The standard softmax function is often enhanced by incorporating additional marginal penalization, which imposes constraints on the feature distributions of different classes. Various methods exist for introducing this penalization. For example, the ArcFace loss matches well with the L2 feature normalization and is formulated by adding an angular marginal penalization parameter $m$ to the positive logit:

\begin{equation}
    L_{ArcFace} = \frac{1}{N}\sum_{i} -\log (\frac{e^{s\cos(\theta_{y_i} + m)}}{e^{s\cos(\theta_{y_i} + m)} + \sum_{j\neq y_i} e^{s\cos\theta_{j}}})
\end{equation}


This modification enforces tighter constraints on the feature distribution of the ground truth class for each sample, thereby increasing the margins between the feature distributions of different classes. The widespread adoption of this formulation stems from its ability to produce highly discriminative and compact class features, while also ensuring robust convergence \cite{arcface_paper, WebFace260M}. While our approach is adaptable to various forms of marginal penalization, we will focus our discussion on the ArcFace framework for simplicity.

\section{Training Schematics}


To address our target properties, namely the requirement of unambiguous classification, we modify the general training schematics as follows. To account for the dual nature of morph samples, we add an additional classification layer to the common architecture, which receives a distinct set of labels for the morph samples. Both classification layers contribute to the overall loss with distinct losses.  

In this scheme, for morph samples (generated from two original subjects), the first loss is computed using the original label of the first subject, while the second loss is computed using the original label of the second subject. The overall schematic is illustrated in Fig. \ref{fig:morphguard_schematic}.

The modified formulation of the loss function (See Eq. \ref{eq:loss_mg}) includes two components which are computed identically for the Bona Fide samples and differently for Morphs.
This creates the driver for the separation of morphs and their original subjetcs if the deep latent feature space. 

\begin{equation}
\begin{aligned}
    L_{MG} = \frac{1}{N}\sum_{i} -(\log (\frac{e^{s\cos(\theta_{\dot{y_i}} + m_{i})}}{e^{s\cos(\theta_{\dot{y_i}} + m_{i})} + \sum_{j\neq \dot{y_i}} e^{s\cos\theta_{j}}}) + \\ 
    +\log (\frac{e^{s\cos(\theta_{\ddot{y_i}} + m_{i})}}{e^{s\cos(\theta_{\ddot{y_i}} + m_{i})} + \sum_{j\neq \ddot{y_i}} e^{s\cos\theta_{j}}})),
\end{aligned}
\label{eq:loss_mg}
\end{equation}
where $m_{i} = m_{M} = m_{BF} + m_{MG}$ for the Morph samples and $m_{i} = m_{BF}$ ($m_{MG}$ and $m_{BF}$ are the constant margin values for Morph and Bona Fide samples). 


Our strategy focuses on intentionally adjusting the balance between morphs and bona fide samples within the feature domain. The parameter $m_{MG}$ can take both positive and negative values, where positive values encourage a stronger tightening of morphs with their assigned ground truth class, while negative values allow for a softer distribution of morph samples compared to bona fide samples of the same class. Both scenarios will be explored in further experiments.



The proposed strategy can indeed be universally adopted to other commonly used marginal softmax-based approaches and integrated into existing pipelines for training face recognition networks.  To achieve this, one needs to duplicate the last classification layer and define a multitask problem where each input is assigned two similar ground truth label outputs. For original images, these outputs remain identical. For face morphs, these outputs correspond to the labels of the original subjects. Further training can then proceed with face morphs defined within the same domain as the original identity subjects.

\subsection{Feature Distribution}

\begin{figure}[htbp]
\centering

\includegraphics[width=0.7\linewidth]{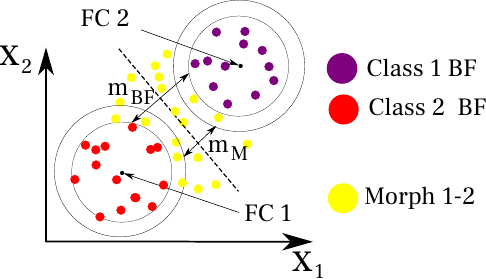}

\caption{Desired feature distribution of proposed method. $m_{M}$ and $m_{BF}$ indicate the margin between the classes for Morphs and Bona Fides.}
\label{fig:MG_feature_distribution_morphguard}
\end{figure}

It is important to analyze the impact of our strategy on deep feature distribution. A simplified example is illustrated in Fig. \ref{fig:MG_feature_distribution_morphguard}. This figure demonstrates the distribution of two original classes and their corresponding morphs within a two-dimensional feature space. Introducing margin parameters $m_{i}$ between classes results in the concentration of samples around their respective feature centers. However, the margins differ for morphs ( $m_{M}$ ) and bona fide samples ( $m_{BF}$ ), intentionally creating an imbalance in their feature distributions. 

Additionally, our strategy employs two distinct classification layers that assign different ground truth labels to morph samples. For instance, as depicted in Fig. \ref{fig:MG_feature_distribution_morphguard}, morphs 1 and 2 are directed toward separate feature regions, effectively differentiating them from their original classes. This separation is further influenced by the imbalance between  $m_{M}$ and  $m_{BF}$, which can positively contribute to the separation of Morphs and Bona Fides. These margins essentially define the target distance between features of ground truth classes (in the case of ArcFace, the feature distribution is affected in a radial manner).

\section{Data Curation}



Since our methodology aligns with key aspects of existing morphing attack detection (MAD) approach \cite{MorDeephy}, it is essential to adopt similar data curation strategies. Specifically, our method requires a large labeled face dataset paired with corresponding morphed images for the same set of identities. Following the principles outlined in \cite{MorDeephy}, we utilize a filtered wild dataset as the source of bona fide images.  

In this work, the wild dataset undergoes a filtering process based on its \textit{suitability for face morphing}. This selection strategy relies on thresholding quality metrics, as proposed in \cite{MorDeephy, quality_driven}, to ensure that only high-quality images are included. The quality assessment is performed using a combination of metrics: Blur \cite{variance_of_laplacian}, FaceQNet \cite{faceQnet}, BRISQUE \cite{brisque}, Face Illumination \cite{fiiqa}, and Pose \cite{pose}. These metrics collectively evaluate different aspects of image quality, including sharpness and natural clarity (Blur, BRISQUE), suitability for ID document authentication (FaceQNet), and acquisition conditions such as lighting and head positioning (Illumination, Pose).  

For the wild dataset, we utilize VGGFace2 \cite{VGGface2}, which contains approximately 3 million images spanning ~9,000 identities, with an average of 360 samples per class. Compared to other widely used datasets such as CASIA-WebFace \cite{casia_webface}, MS-Celeb-1M \cite{ms_celeb_face, MS-Celeb-1M_CleanList}, and Glint360K \cite{Partial_fc}, VGGFace2 provides a larger number of samples per identity. This characteristic ensures that a sufficient number of images remain available even after filtering. As a result of this filtering process, the final curated dataset comprises approximately 500,000 images.

\subsection{Morph Dataset}

Our training approach incorporates face morphs into the learning process and require a specific dataset of morphed images corresponding to bona fide identities. However, the identity pairing strategy for generating these morphs cannot be arbitrary. A fundamental requirement for effective training is to ensure unambiguous class labeling of morphed images to prevent ambiguity in classification.  

When a face morph is generated from two samples within the original dataset, it inherently possesses a dual identity. However, the assignment of these identities to specific labels, denoted as $\dot{y}$ and $\ddot{y}$ (see Fig.~\ref{fig:morphguard_schematic}), is not inherently defined. Consequently, if images are randomly paired without constraints, classification inconsistencies arise due to the ambiguous labeling of the morphed samples.  

To address this, we introduce a structured pairing strategy. First, the total set of identities is divided into two disjoint subsets, each assigned to one of the two branches in our training framework (see Fig.~\ref{fig:morphguard_schematic}). Morphed images are then generated by pairing samples exclusively between identities from different subsets. Each morphed image is labeled according to the subset from which its source identities originate, ensuring consistent classification within each network branch. This labeling scheme is applied uniformly to morphed and bona fide images.  

The idea of this separation is to ensure that the morphed images retain a clear identity association between the two branches of the network. For example, given a dataset containing four identities: $[A, B, C, D]$, we divide them into two disjoint sets: $[A, B]$ (assigned to the \textit{First} network) and $[C, D]$ (assigned to the \textit{Second} network). Morphs are then created by pairing identities across these subsets, yielding combinations such as $[AC]$, $[BC]$, $[AD]$, and $[BD]$, each of which can be distinctly classified. In contrast, within-subset pairings, such as $[AB]$ or $[CD]$, would introduce classification ambiguity and are therefore avoided.  

This structured identity separation is maintained across different morphing methods, with variations in the specific pairing sampling strategies. By ensuring controlled pairing protocols while preserving the integrity of the data set, our approach mitigates classification confusion and enhances the robustness of the learning process.

\subsection{Selfmorphing}




Fully automated face morphing techniques often introduce visible artifacts, such as blending patterns or specific statistical features, into the generated images. Without proper regularization, a detection model may inadvertently learn to recognize these artifacts rather than the underlying morphing-specific defects, leading to an unrealistic bias. To mitigate this artifact-driven bias, we incorporate \textit{selfmorphs} — morphed images generated from two different instances of the same identity.  

The use of selfmorphing is not a novel concept. Borghi \textit{et al.} \cite{morphing_artifacts_retouching} previously used it to introduce artificial artifacts, which were later used to train a GAN model for their removal. The other work \cite{MorDeephy} leveraged selfmorphing to minimize the reliance on artifacts in morphing attack detection.  

In our approach, we assume that the core features of facial identity remain preserved after selfmorphing. In the schematic of the proposed method  (Fig.~\ref{fig:morphguard_schematic}), selfmorphs are treated as bona fide samples, ensuring a more balanced training process.  

The pairing strategy for generating selfmorphs follows a simple protocol — images within the same identity class are randomly paired across the dataset to create morphed samples.

\subsection{Overall Training Dataset}


Using the defined pairing protocols, we construct the training dataset of face morphs by employing multiple morphing techniques. Specifically, we generate morphs using a custom landmark-based morphing approach (with a blending coefficient of $0.5$), the StyleGAN-based method \cite{styleGAN}, and a Diffusion Autoencoder-based approach \cite{MorDIFF, diffae}.  

For each morphing technique, we produce a dataset containing $\sim$ 250K morphed images, including both standard face morphs and selfmorphs. When combined with the original bona fide images, these datasets form the complete training set used in our experiments.

\section{Experiments}

To demonstrate the effectiveness of our approach, we conduct a series of experiments and analyze their results in this section. However, before presenting the experimental findings, we first discuss the benchmarking methodology.

\subsection{Benchmarking}



In a practical face morphing attack scenario, the goal is to successfully deceive face recognition systems by matching the morphed image to both original identities. To evaluate the robustness of biometric systems against such attacks, a dedicated metric, the Mated Morph Presentation Match Rate (MMPMR), was introduced \cite{Biometric_Systems_under_Morphing_Attacks}.  

The MMPMR is defined as follows:

\begin{equation}
\label{eq:mmpmr}
\begin{aligned}
MMPMR(\tau ) = {\color{white} \cdot\cdot\cdot\cdot\cdot\cdot\cdot\cdot\cdot\cdot } \\  \frac{1}{M} \cdot \sum_{m=1}^{M} \left \{ \left ( \begin{matrix}
min\\ 
n =1,...,N_{m}
\end{matrix} 
S_{m}^{n} \right )>\tau \right \}.
\end{aligned}
\end{equation}



Here, $\tau$ represents the decision threshold, $S_{m}^{n}$ denotes the similarity score of the n-th subject within the morph $m$, $M$ is the total number of morphed images, and $N_{m}$ refers to the number of subjects contributing to the morph $m$ (typically two in practical scenarios).  

Additionally, a more generalized metric that evaluates both face recognition performance and robustness against morphing attacks has been introduced. The Relative Morph Match Rate (RMMR) is formulated as a balance between MMPMR and the True Match Rate (TMR):

\begin{equation}
\label{eq:rmmr}
\begin{aligned}
RMMR(\tau) = 1+(MMPMR(\tau)-TMR(\tau))=\\= MMPMR(\tau)+FNMR(\tau). {\color{white} \cdot\cdot\cdot \cdot\cdot\cdot \cdot\cdot\cdot}
\end{aligned}
\end{equation}


In this study, benchmarking using the defined metrics is conducted with the MorFacing utilities \cite{morfacing}. This benchmarking framework incorporates all essential protocols, including the morphing pairing protocol (for generating morphed images), 1-1 verification (for evaluating face recognition performance), and MMPMR calculation (for assessing robustness against morphing attacks). It enables a comprehensive comparison of different face recognition approaches across various evaluation metrics, such as False Match Rate (FMR), False Non-Match Rate (FNMR), and morphing robustness metrics.  

The benchmark is based on a dataset curated by Guerra et al. \cite{GuerraMG23}, specifically designed to analyze compliance with ICAO standards. This dataset comprises a diverse set of samples reflecting a wide range of noncompliance cases, making it well-suited for studying face morphing attack vulnerabilities. Additionally, the dataset includes images captured using different devices (smartphones and DSLR cameras), allowing for a realistic evaluation scenario where images in verification pairs are acquired under varying conditions.  

Our evaluation follows the morphing pairing protocol outlined in \cite{morfacing}. We use  face morphs generated with multiple techniques, including the Landmark-Based OpenCV method \cite{learn_opencv_morpher} (two protocols: \textit{LDM Raw} - straightforward warping and 
blending in image domain and \textit{LDM}, with Seamless Cloning used for background restoration), StyleGAN (\textit{STG})\cite{styleGAN}, Diffusion Autoencoders (\textit{DiffAE})\cite{MorDIFF}, and Neural Implicit Morphing of Face Images (two protocols: \textit{NIM S.Mix} - warping with seamless mix blending, \textit{NIM DiffAE} - blending with use of DiffAE) \cite{Schardong_2024_CVPR}.

\subsection{Morph Margin Balance}
\label{section:margin_balance}

We conducted a series of experiments using the proposed methodology and the selected dataset. In all experiments, the \textit{ConvNeXt Tiny} architecture was trained on images resized to \(224 \times 224\) pixels. \textit{ConvNeXt} is a modern convolutional neural network (CNN) architecture that incorporates design elements inspired by transformer-based models while maintaining the efficiency of traditional CNNs. It has demonstrated competitive performance on various computer vision tasks, making it a strong alternative to vision transformers. We chose \textit{ConvNeXt Tiny} due to its advanced hierarchical design, improved training stability, and superior feature extraction capabilities, which align well with the requirements of classical face recognition tasks.


Our initial experiments focused on analyzing the balance of components within the complex margin \( m_{i} \). Specifically, the values of these margin components influence the distribution of morph samples in relation to their assigned ground-truth classes. This effect determines the degree of penalization applied to morph samples: increasing the margin can push morph samples further apart or, conversely, allow them to overlap more with bona fide samples. Both scenarios introduce imbalances in the feature space, potentially affecting the robustness to face morphing attacks. Our goal here is to identify a balance between the margins of the loss function.  

To achieve this, we conducted multiple training sessions with different balance settings in our loss function. We set a baseline margin for bona fide samples at \( m_{BF} = 0.5 \) and  varied the additive margin component for morph samples, $ m_{MG} $. The training process took 10 epochs and Stochastic Gradient Descent (SGD) was employed as the optimization method, with a learning rate that followed a linear decay schedule from 0.001 to 0.00001 over the course of training.



\begin{figure*}[htbp]
\centering

\includegraphics[width=0.32\linewidth]{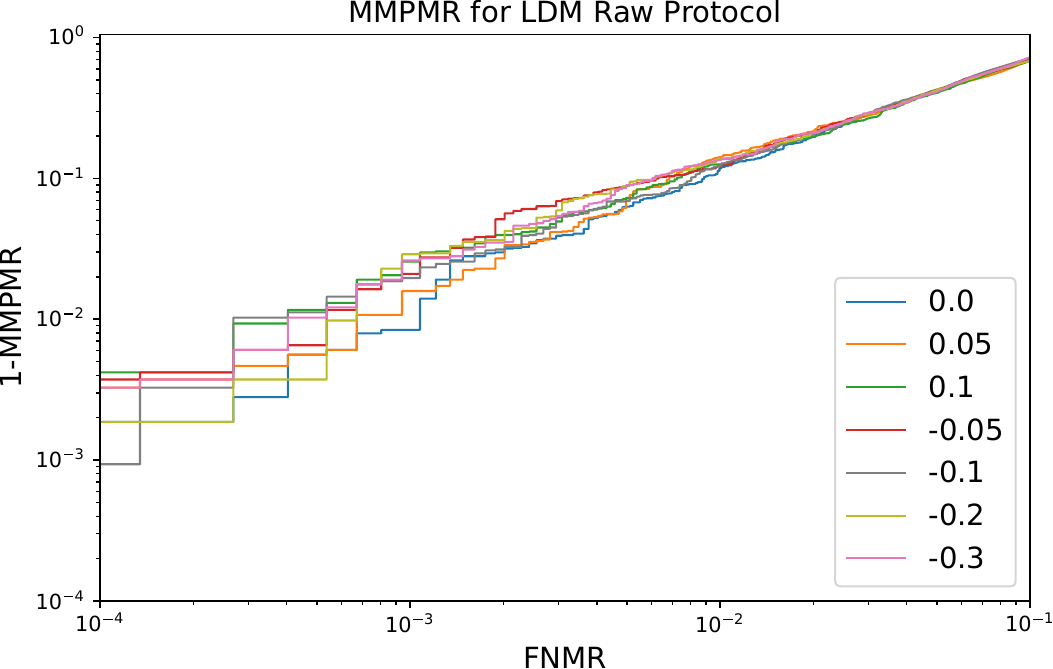}
\includegraphics[width=0.32\linewidth]{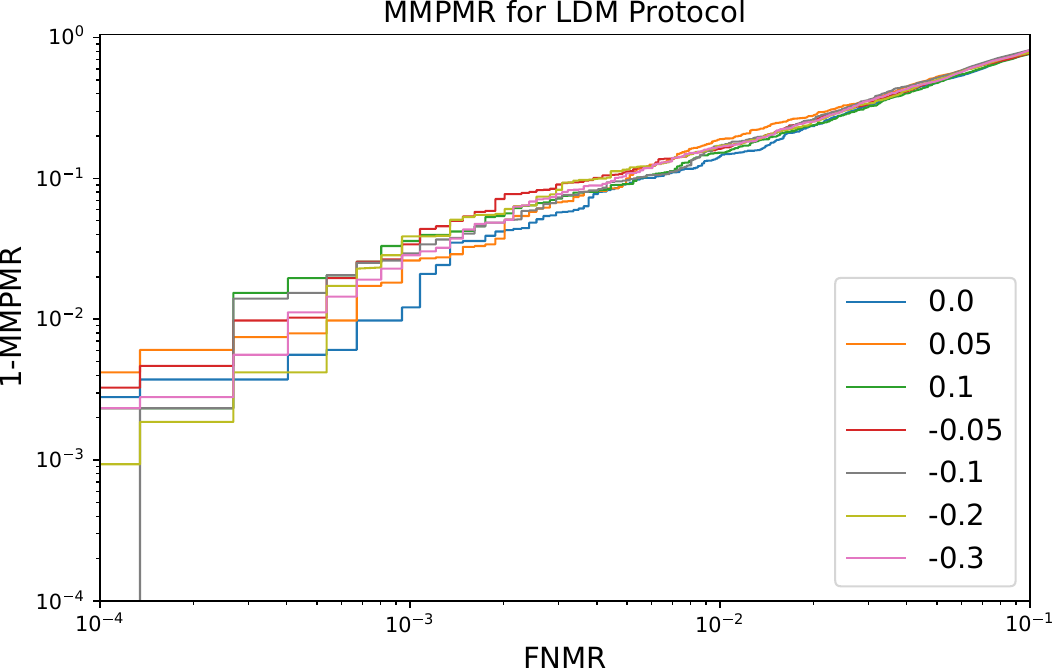}
\includegraphics[width=0.32\linewidth]{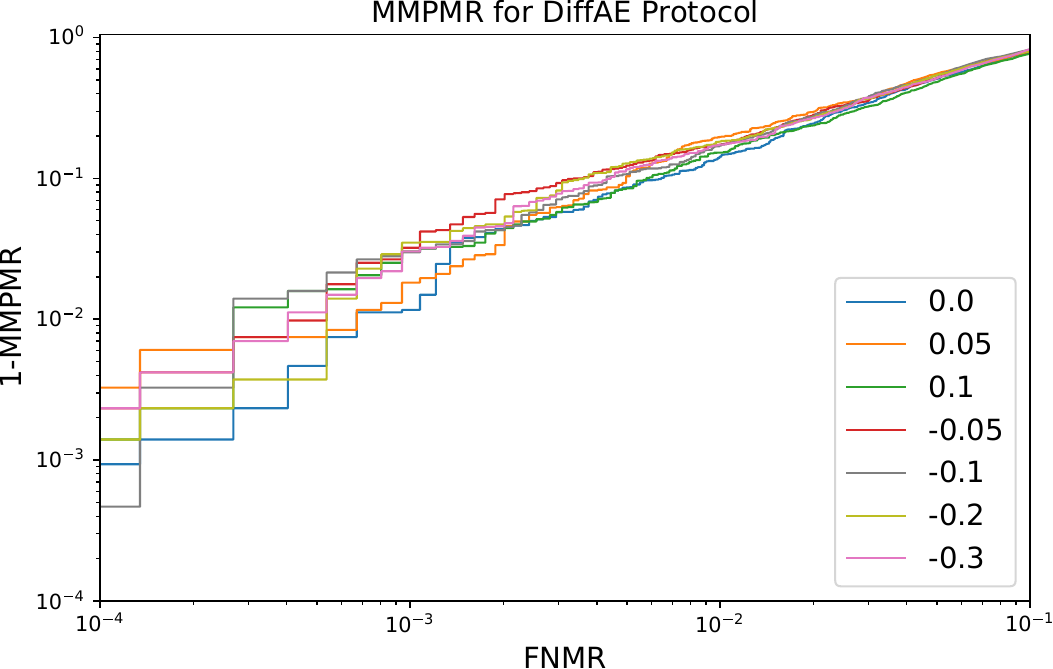}

 a \hspace{150pt}   b \hspace{150pt} c

\vspace{5pt}
\includegraphics[width=0.32\linewidth]{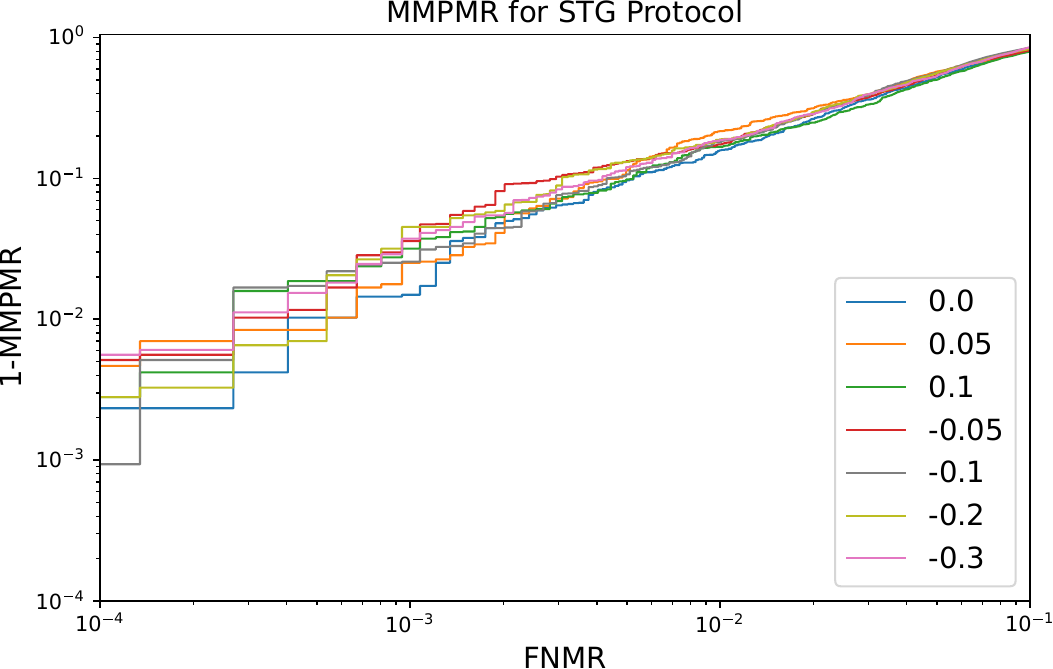}
\includegraphics[width=0.32\linewidth]{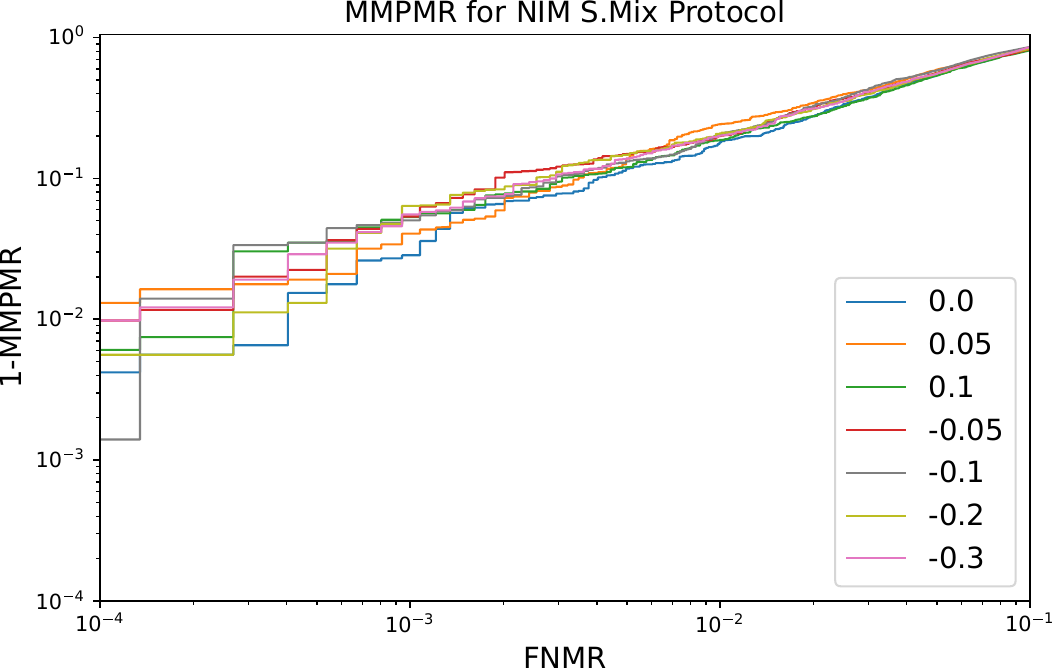}
\includegraphics[width=0.32\linewidth]{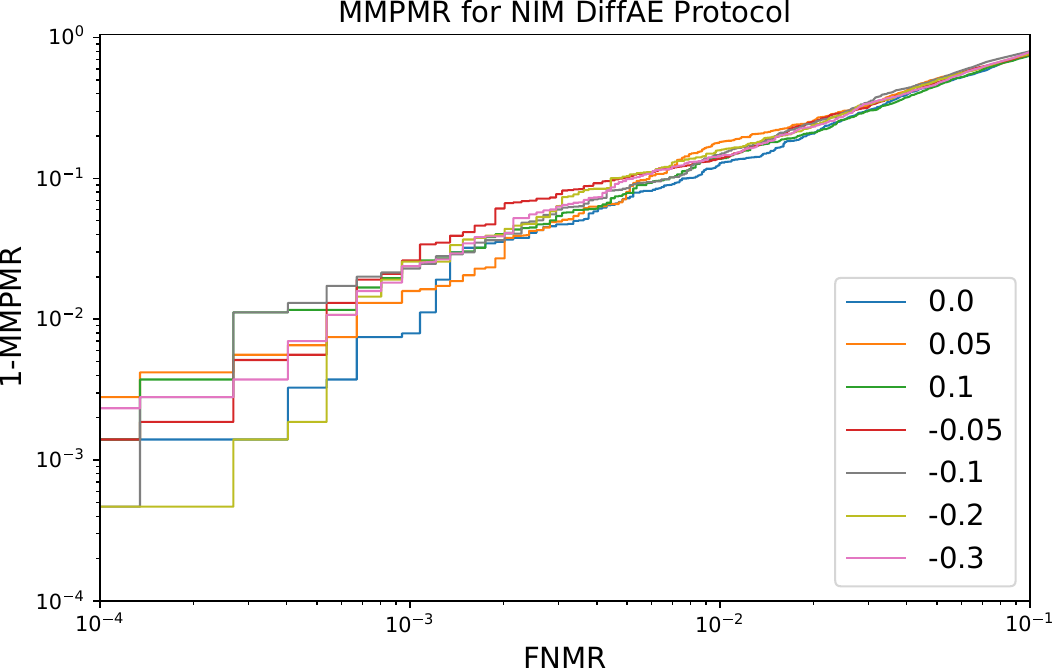}

 d \hspace{150pt}   e \hspace{150pt} f

\caption{MMPMR for different models and protocols as a dependency from FNMR. For convenience vertical axis is plotted as 1-MMPMR  (higher values \textrightarrow better robustness).  a - LDM Raw Protocol, b - LDM protocol; c - DiffAE Protocol, d - STG Protocol, e - NIM S. Mix Protocol, f - NIM DiffAE Protocol.}
\label{fig:MG_MMPMR}
\end{figure*}

The key performance metrics for evaluating the robustness of our approach against morphing attacks are presented in terms of the dependency of MMPMR (Mated Morph Presentation Match Rate) on FNMR (False Non-Match Rate) values (see Fig.~\ref{fig:MG_MMPMR}). Specific results for $MMPMR @ FNMR = [0.01, 0.001] $ are summarized in Table~\ref{tab:MG_MMPMR_results}.  

\begin{figure}[htbp]
\centering

\includegraphics[width=0.9\linewidth]{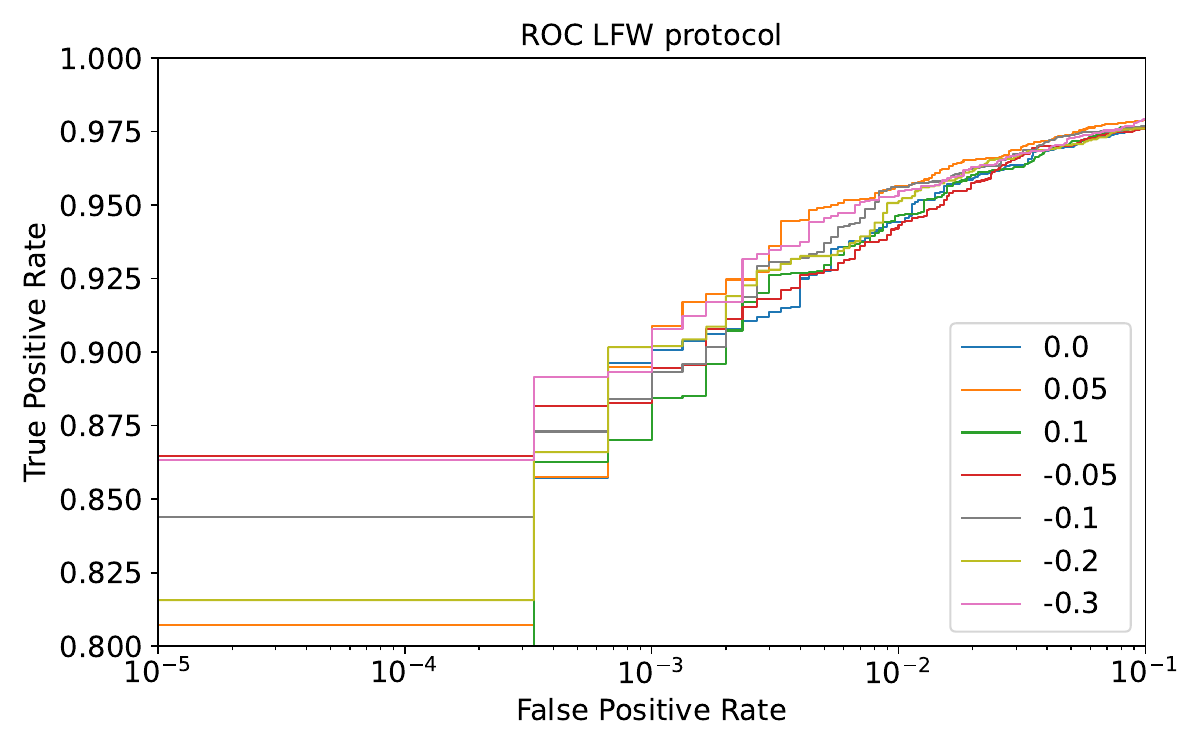}

 a 

\vspace{5pt}
\includegraphics[width=0.9\linewidth]{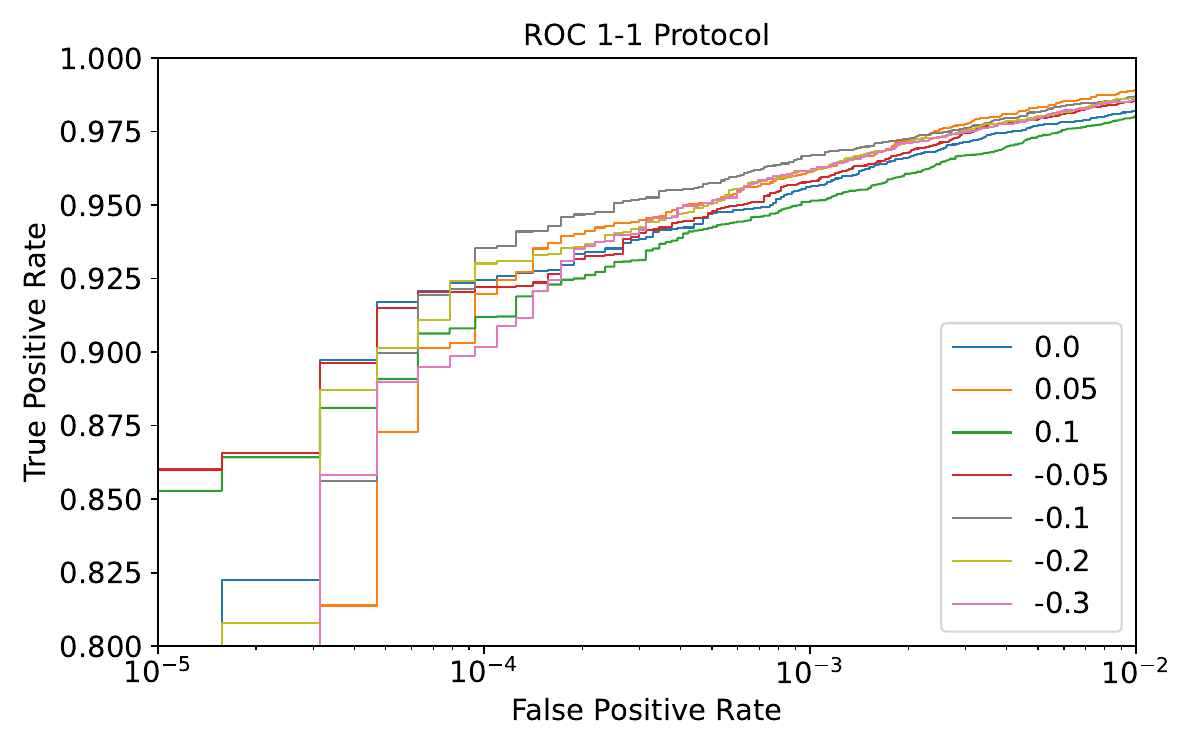}

 b

\caption{ROC curves for the 1-1 verification performance on LFW (a) and 1-1 MorFacing protocol(b)}
\label{fig:MG_11_protocols}
\end{figure}

Additionally, we provide collateral results for 1-1 verification performance using the 1-1 MorFacing Protocol, along with supplementary performance evaluation on the LFW benchmark. These results are visualized in Fig.~\ref{fig:MG_11_protocols} in the form of a ROC curve and are further detailed in Table~\ref{tab:MG_11_table_results} as  $FNMR@FMR = [0.001, 0.0001]$.  

Our evaluation considers three different settings for $m_{MG}$, such as:  positive $m_{MG}$, which increases the penalization of morph samples; $m_{MG} = 0$ (baseline case), corresponding to the standard ArcFace model without any additional imbalance; Negative $m_{MG}$, which reduces the penalization of morph samples.  

Based on these results, we derive several key observations regarding the effectiveness of our proposed strategy in enhancing robustness against morphing attacks.




Based on the overall analysis of Fig.~\ref{fig:MG_MMPMR} and Table~\ref{tab:MG_MMPMR_results}, we conclude that introducing any form of margin imbalance, whether positive or negative, has a beneficial impact on the robustness against morphing attacks. However, the extent of this improvement depends on the specific margin values used. 

Conceptually, this aligns with the feature distribution observed in Fig.\ref{fig:MG_feature_distribution_morphguard}, where morph samples remain more distant from their assigned ground truth class center. This separation reduces their likelihood of being misclassified, thereby enhancing the system’s robustness to morphing attacks.

At the same time, excessively high absolute values of $ m_{MG} $ diminish this positive effect. In our experiments, a high positive  $m_{MG}$  often resulted in convergence issues, similar to the well-documented instability of ArcFace loss when using large margin values. Conversely, for negative $m_{MG}$, increasing its absolute value led to a decrease in the beneficial impact on robustness, suggesting that a balanced approach is necessary to achieve optimal performance.

We also examined additional 1-1 verification results from the 1-1 MorFacing protocol and the LFW \cite{LFW_dataset} benchmark. Notably, our strategy had a positive impact on overall face verification performance, attributed to the more sophisticated approach to sample labeling. 
This effect is more noticeable in the 1-1 MorFacing protocol, which consists of good quality frontal images, whereas the performance on wild images remained largely unchanged in cases with a disbalanced margin.

\begin{figure*}[htbp]
\centering

\includegraphics[width=0.32\linewidth]{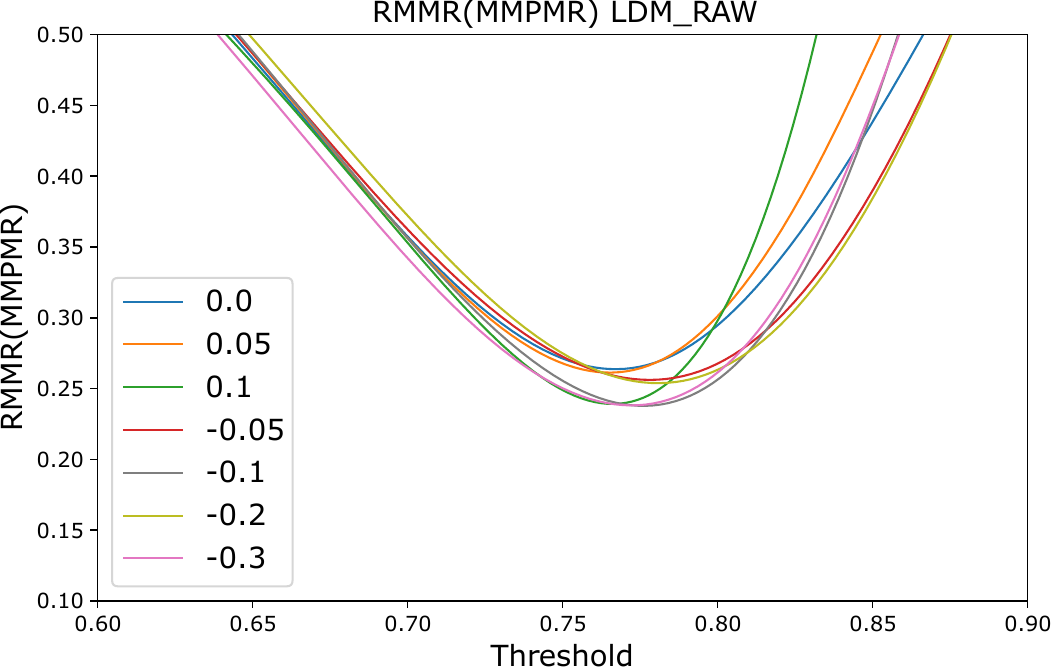}
\includegraphics[width=0.32\linewidth]{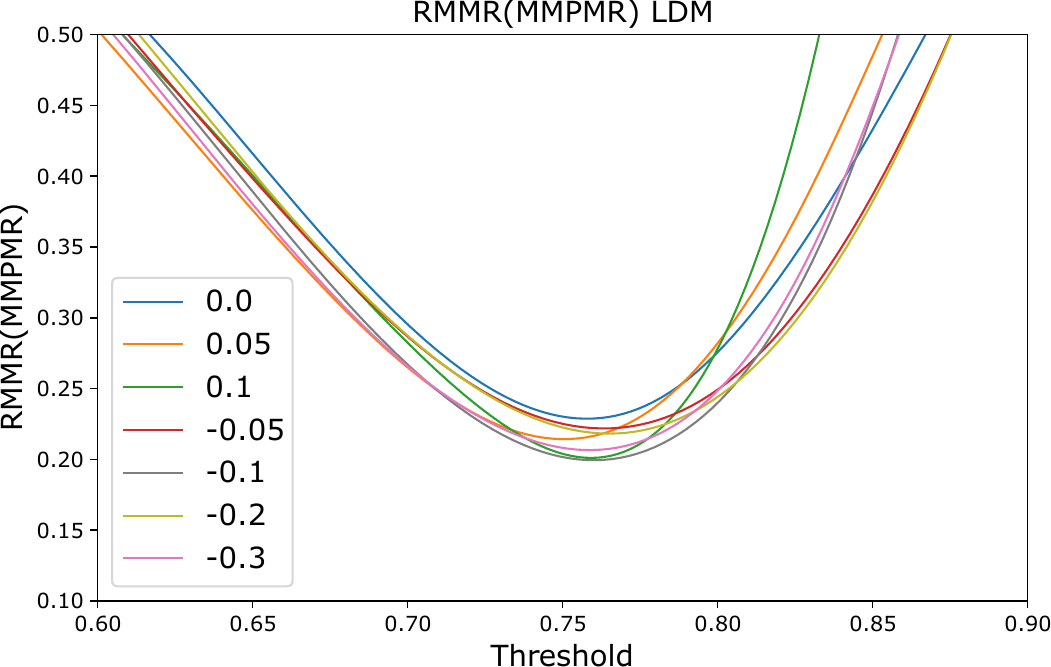}
\includegraphics[width=0.32\linewidth]{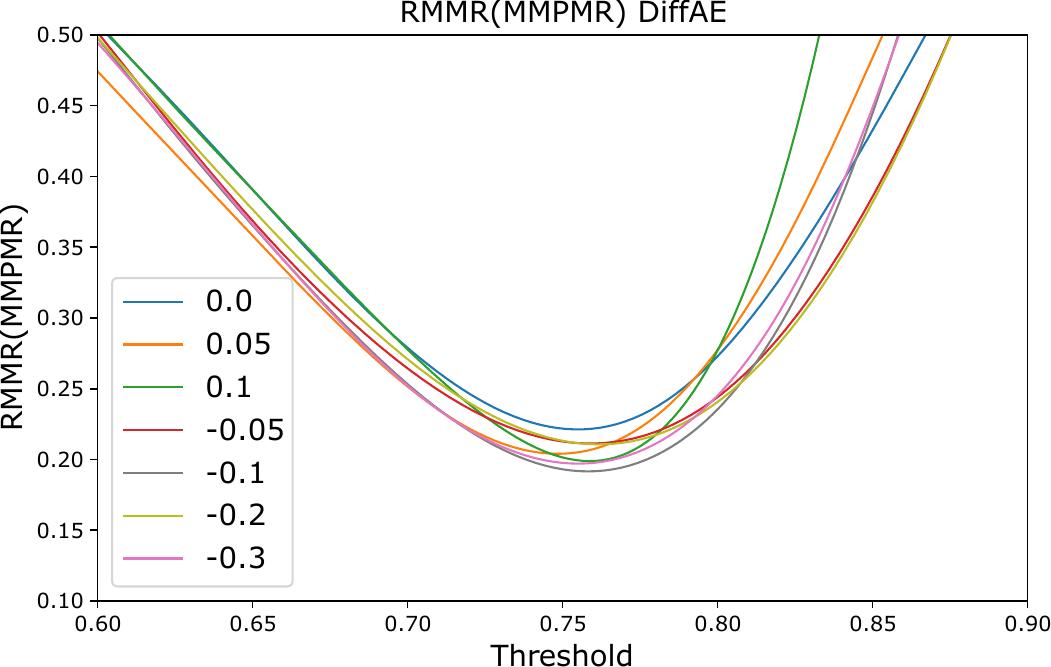}

 a \hspace{150pt}   b \hspace{150pt} c

\vspace{5pt}
\includegraphics[width=0.32\linewidth]{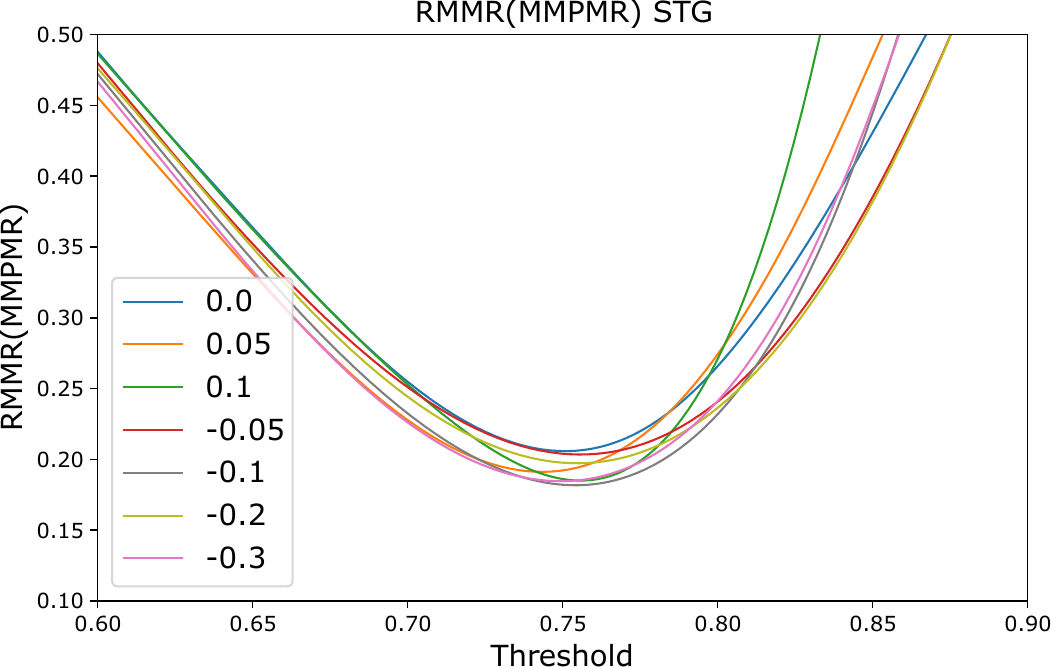}
\includegraphics[width=0.32\linewidth]{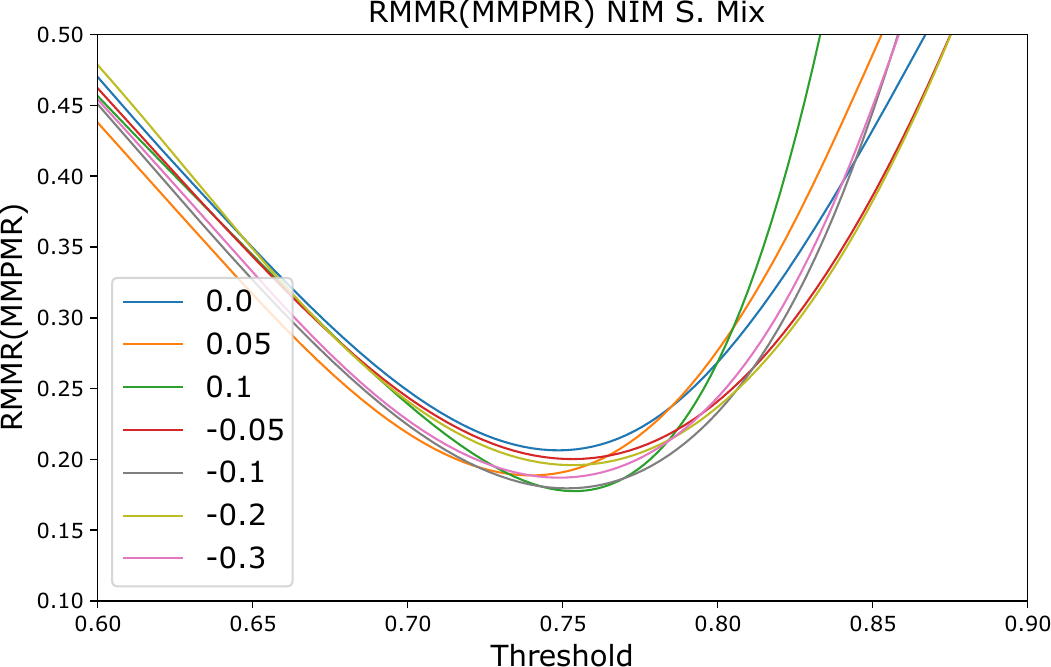}
\includegraphics[width=0.32\linewidth]{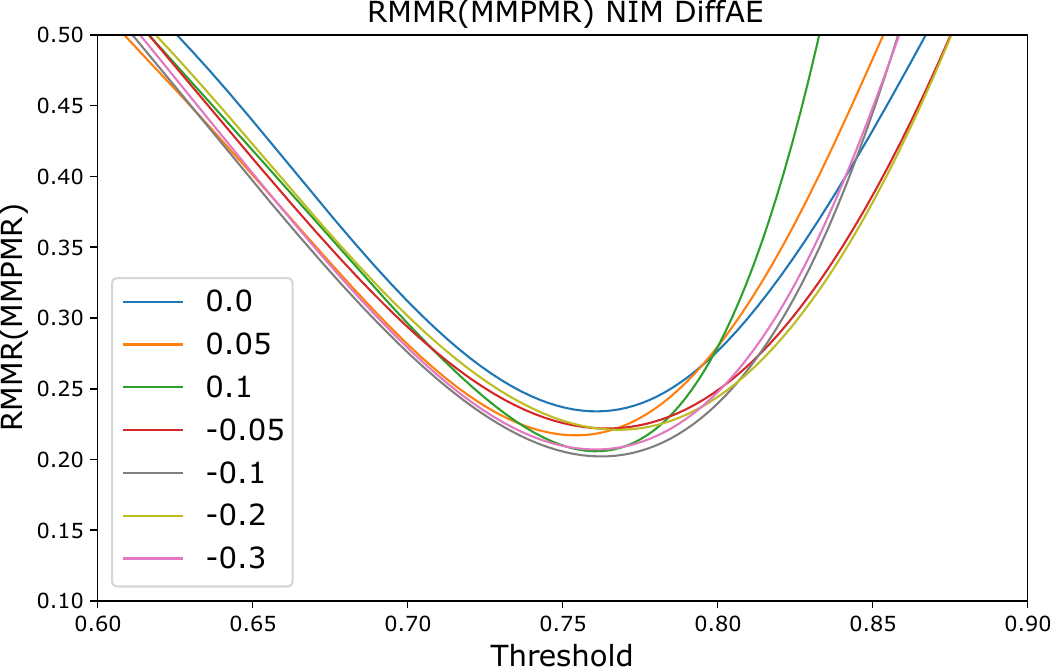}

 d \hspace{150pt}   e \hspace{150pt} f

\caption{RMMR for different models and protocols as a dependency from similarity threshold.  a - LDM Raw Protocol, b - LDM protocol; c - DiffAE Protocol, d - STG Protocol, e - NIM S. Mix Protocol, f - NIM DiffAE Protocol.}
\label{fig:MG_RMMR}
\end{figure*}

We also evaluated the trade-off between face recognition performance and robustness against face morphing attacks using the Relative Morph Match Rate (RMMR) metric for the trained models (see Fig. \ref{fig:MG_RMMR} and Table \ref{tab:MG_RMMR_results}). This trade-off is assessed as the minimum RMMR value within the available threshold range, as presented in Table \ref{tab:MG_RMMR_results}. The results further highlight the positive impact of margin imbalance, providing valuable insights into optimal parameter selection, with the most effective margin disbalance in our experiments found to be $m_{MG} = -0.1$.

Based on the analysis of all results regarding robustness to morphing attacks and face recognition performance, we conclude that the value  $m_{MG} = -0.1$ gives the most optimal performance, achieving the best margin balance among the tested values.

\begin{table}[]
\caption{1-1 Verification performance as FNMR@FMR = [0.001, 0.0001] for models with different values of $m_{MG}$ ($m_{BF} = 0.5$) on two protocols - 1-1 MorFacing(MF) and LFW.}
\vspace{5px}
\resizebox{0.49\textwidth}{!}{
\begin{tabular}{|c|ll|ll|}
\hline
\multirow{2}{*}{\begin{tabular}[c]{@{}c@{}}\textit{1-1 Verification}\\ \textit{Performance $m_{MG}=$}\end{tabular}} & \multicolumn{2}{c|}{\textit{1-1 MF; FMR =}}                & \multicolumn{2}{c|}{\textit{LFW; FMR =}}     \\ \cline{2-5} 
                                                                                           & \multicolumn{1}{c|}{\textit{0.001}} & \multicolumn{1}{c|}{\textit{0.0001}} & \multicolumn{1}{c|}{\textit{0.001}} & \textit{0.0001} \\ \hline
0.1                                                                                        & \multicolumn{1}{l|}{0.187}      &       0.607                   & \multicolumn{1}{l|}{0.115}      &    0.396   \\ \hline
0.05                                                                                       & \multicolumn{1}{l|}{0.177}      &          0.378                  & \multicolumn{1}{l|}{\textbf{0.090}}      &   0.192     \\ \hline
0.0                                                                                         & \multicolumn{1}{l|}{0.315}      &           0.661                  & \multicolumn{1}{l|}{0.099}      &   0.208     \\ \hline
-0.05                                                                                       & \multicolumn{1}{l|}{0.138}      &          \textbf{0.406}                  & \multicolumn{1}{l|}{0.105}      &    \textbf{0.135}   \\ \hline
-0.1                                                                                        & \multicolumn{1}{l|}{0.157}      &     0.410                        & \multicolumn{1}{l|}{0.106}      &   0.156     \\ \hline
-0.2                                                                                        & \multicolumn{1}{l|}{\textbf{0.120}}      &         0.515                    & \multicolumn{1}{l|}{0.097}      &    0.184    \\ \hline
-0.3                                                                                        & \multicolumn{1}{l|}{0.143}      &            0.464                 & \multicolumn{1}{l|}{0.092}      &   0.136     \\ \hline
\end{tabular}}
\label{tab:MG_11_table_results}
\end{table}

\begin{table}[]
\caption{MMPMR values for the models trained for estimation of the margin balance. Margin $m_{BF} = 0.5$}
\vspace{5px}
\resizebox{0.49\textwidth}{!}{\begin{tabular}{|cl|lllllll|}
\hline
\multicolumn{2}{|c|}{\multirow{2}{*}{\begin{tabular}[c]{@{}c@{}}\textit{Margin Balance} \textit{vs.}\\ \textit{Protocols}\end{tabular}}} & \multicolumn{7}{c|}{\textit{Margin} $m_{MG} =$  }                                                                                                                                        \\ \cline{3-9} 
\multicolumn{2}{|c|}{}                                                                                        & \multicolumn{1}{|c|}{\textit{0.1}} & \multicolumn{1}{|c|}{\textit{0.05}} & \multicolumn{1}{|c|}{\textit{0.0}} & \multicolumn{1}{|c|}{\textit{-0.05}} & \multicolumn{1}{|c|}{\textit{-0.1}} & \multicolumn{1}{|c|}{\textit{-0.2}} & \textit{-0.3} \\ \hline
\multicolumn{1}{|c|}{\multirow{6}{*}{\rotatebox[origin=c]{90}{FNMR = 0.01}}}                           & LDM Raw                              & \multicolumn{1}{l|}{0.874}    & \multicolumn{1}{l|}{\textbf{0.859}}     & \multicolumn{1}{l|}{0.886}    & \multicolumn{1}{l|}{0.877}      & \multicolumn{1}{l|}{0.877}    & \multicolumn{1}{l|}{0.863}    &  0.865   \\ \cline{2-9} 
\multicolumn{1}{|c|}{}                                                 & LDM                                  & \multicolumn{1}{l|}{0.847}    & \multicolumn{1}{l|}{\textbf{0.811}}     & \multicolumn{1}{l|}{0.859}    & \multicolumn{1}{l|}{0.837}      & \multicolumn{1}{l|}{0.830}    & \multicolumn{1}{l|}{0.830}    & 0.832    \\ \cline{2-9} 
\multicolumn{1}{|c|}{}                                                 & DiffAE                               & \multicolumn{1}{l|}{0.847}    & \multicolumn{1}{l|}{\textbf{0.803}}     & \multicolumn{1}{l|}{0.860}    & \multicolumn{1}{l|}{0.826}      & \multicolumn{1}{l|}{0.830}    & \multicolumn{1}{l|}{0.818}    & 0.827    \\ \cline{2-9} 
\multicolumn{1}{|c|}{}                                                 & STG                                  & \multicolumn{1}{l|}{0.832}    & \multicolumn{1}{l|}{\textbf{0.784}}     & \multicolumn{1}{l|}{0.843}    & \multicolumn{1}{l|}{0.825}      & \multicolumn{1}{l|}{0.819}    & \multicolumn{1}{l|}{0.811}    &  0.814   \\ \cline{2-9} 
\multicolumn{1}{|c|}{}                                                 & NIM S.Mix                            & \multicolumn{1}{l|}{0.813}    & \multicolumn{1}{l|}{\textbf{0.758}}     & \multicolumn{1}{l|}{0.824}    & \multicolumn{1}{l|}{0.800}      & \multicolumn{1}{l|}{0.799}    & \multicolumn{1}{l|}{0.792}    & 0.800    \\ \cline{2-9} 
\multicolumn{1}{|c|}{}                                                 & NIM DiffAE                           & \multicolumn{1}{l|}{0.861}    & \multicolumn{1}{l|}{\textbf{0.820}}     & \multicolumn{1}{l|}{0.876}    & \multicolumn{1}{l|}{0.861}      & \multicolumn{1}{l|}{0.853}    & \multicolumn{1}{l|}{0.841}    & 0.855   \\ \hline\hline
\multicolumn{1}{|c|}{\multirow{6}{*}{\rotatebox[origin=c]{90}{FNMR = 0.001}}}                           & LDM Raw                              & \multicolumn{1}{l|}{0.979}    & \multicolumn{1}{l|}{0.989}     & \multicolumn{1}{l|}{0.991}    & \multicolumn{1}{l|}{0.981}      & \multicolumn{1}{l|}{0.981}    & \multicolumn{1}{l|}{\textbf{0.977}}    &   0.980  \\ \cline{2-9} 
\multicolumn{1}{|c|}{}                                                 & LDM                                  & \multicolumn{1}{l|}{\textbf{0.966}}    & \multicolumn{1}{l|}{0.981}     & \multicolumn{1}{l|}{0.990}    & \multicolumn{1}{l|}{0.973}      & \multicolumn{1}{l|}{0.973}    & \multicolumn{1}{l|}{0.971}    & 0.977    \\ \cline{2-9} 
\multicolumn{1}{|c|}{}                                                 & DiffAE                               & \multicolumn{1}{l|}{0.974}    & \multicolumn{1}{l|}{0.986}     & \multicolumn{1}{l|}{0.988}    & \multicolumn{1}{l|}{0.973}      & \multicolumn{1}{l|}{0.972}    & \multicolumn{1}{l|}{\textbf{0.971}}    &   0.978  \\ \cline{2-9} 
\multicolumn{1}{|c|}{}                                                 & STG                                  & \multicolumn{1}{l|}{0.972}    & \multicolumn{1}{l|}{0.982}     & \multicolumn{1}{l|}{0.985}    & \multicolumn{1}{l|}{0.970}      & \multicolumn{1}{l|}{0.974}    & \multicolumn{1}{l|}{\textbf{0.968}}    & 0.971    \\ \cline{2-9} 
\multicolumn{1}{|c|}{}                                                 & NIM S.Mix                            & \multicolumn{1}{l|}{\textbf{0.949}}    & \multicolumn{1}{l|}{0.965}     & \multicolumn{1}{l|}{0.972}    & \multicolumn{1}{l|}{0.952}      & \multicolumn{1}{l|}{0.951}    & \multicolumn{1}{l|}{0.952}    & 0.954    \\ \cline{2-9} 
\multicolumn{1}{|c|}{}                                                 & NIM DiffAE                           & \multicolumn{1}{l|}{0.980}    & \multicolumn{1}{l|}{0.986}     & \multicolumn{1}{l|}{0.992}    & \multicolumn{1}{l|}{0.979}      & \multicolumn{1}{l|}{\textbf{0.978}}    & \multicolumn{1}{l|}{0.980}    &  0.982   \\ \hline
\end{tabular}}
\label{tab:MG_MMPMR_results}
\end{table}

\begin{table}[]
\caption{The mimimum values of RMMR for different models in different testing protocols. Margin $m_{BF} = 0.5$}
\vspace{5px}
\resizebox{0.49\textwidth}{!}{\begin{tabular}{|cl|lllllll|}
\hline
\multicolumn{2}{|c|}{\multirow{2}{*}{\begin{tabular}[c]{@{}c@{}}\textit{Minimum of } \\ \textit{RMMR curves}\end{tabular}}} & \multicolumn{7}{c|}{\textit{Margin} $m_{MG} =$  }                                                                                                                                        \\ \cline{3-9} 
\multicolumn{2}{|c|}{}                                                                                        & \multicolumn{1}{|c|}{\textit{0.1}} & \multicolumn{1}{|c|}{\textit{0.05}} & \multicolumn{1}{|c|}{\textit{0.0}} & \multicolumn{1}{|c|}{\textit{-0.05}} & \multicolumn{1}{|c|}{\textit{-0.1}} & \multicolumn{1}{|c|}{\textit{-0.2}} & \textit{-0.3} \\ \hline
\multicolumn{1}{|c|}{\multirow{6}{*}{\rotatebox[origin=c]{90}{Protocols}}}                           & LDM Raw                              & \multicolumn{1}{l|}{0.239}    & \multicolumn{1}{l|}{0.261}     & \multicolumn{1}{l|}{0.263}    & \multicolumn{1}{l|}{0.256}      & \multicolumn{1}{l|}{\textbf{0.237}}    & \multicolumn{1}{l|}{0.253}    &  0.238   \\ \cline{2-9} 
\multicolumn{1}{|c|}{}                                                 & LDM                                  & \multicolumn{1}{l|}{0.201}    & \multicolumn{1}{l|}{0.214}     & \multicolumn{1}{l|}{0.228}    & \multicolumn{1}{l|}{0.221}      & \multicolumn{1}{l|}{\textbf{0.199}}    & \multicolumn{1}{l|}{0.218}    & 0.206    \\ \cline{2-9} 
\multicolumn{1}{|c|}{}                                                 & DiffAE                               & \multicolumn{1}{l|}{0.198}    & \multicolumn{1}{l|}{0.204}     & \multicolumn{1}{l|}{0.221}    & \multicolumn{1}{l|}{0.211}      & \multicolumn{1}{l|}{\textbf{0.191}}    & \multicolumn{1}{l|}{0.210}    & 0.197    \\ \cline{2-9} 
\multicolumn{1}{|c|}{}                                                 & STG                                  & \multicolumn{1}{l|}{0.184}    & \multicolumn{1}{l|}{0.191}     & \multicolumn{1}{l|}{0.205}    & \multicolumn{1}{l|}{0.203}      & \multicolumn{1}{l|}{\textbf{0.181}}    & \multicolumn{1}{l|}{0.197}    &  0.184   \\ \cline{2-9} 
\multicolumn{1}{|c|}{}                                                 & NIM S.Mix                            & \multicolumn{1}{l|}{0.177}    & \multicolumn{1}{l|}{0.188}     & \multicolumn{1}{l|}{0.206}    & \multicolumn{1}{l|}{0.200}      & \multicolumn{1}{l|}{\textbf{0.179}}    & \multicolumn{1}{l|}{0.195}    & 0.187    \\ \cline{2-9} 
\multicolumn{1}{|c|}{}                                                 & NIM DiffAE                           & \multicolumn{1}{l|}{0.205}    & \multicolumn{1}{l|}{0.217}     & \multicolumn{1}{l|}{0.233}    & \multicolumn{1}{l|}{0.221}      & \multicolumn{1}{l|}{\textbf{0.202}}    & \multicolumn{1}{l|}{0.220}    & 0.207   \\ \hline
\end{tabular}}
\label{tab:MG_RMMR_results}
\end{table}

\subsection{Pretrained Model Adaptation}
\label{section:model_adaptation}

We conducted an additional experiment to evaluate the application of our strategy to a previously trained face recognition model. Specifically, we designed a two-stage experiment in which, in the first stage, the face recognition network was initially trained on the original dataset containing only bona fide samples. In the second stage, the model was further refined using our proposed strategy, incorporating a dataset augmented with morphs generated from the original images used in the first stage. This approach allows us to assess the impact of our method on improving robustness against morphing attacks while maintaining overall recognition performance.


Here we also employ the \textit{ConvNeXt Tiny} architecture as the backbone, using input images resized to $224 \times 224$ pixels. In the first training stage, the model is trained for 15 epochs only on the original dataset. We increase the number of epochs due to the lower amount of training images in this setting. The learning rate follows a linear decay schedule, decreasing from 0.001 to 0.00001 over the course of training. 

The second stage also spans 10 epochs and is conducted on the full dataset, which includes both original and morphed face images. The learning rate follows a linear decay schedule, decreasing from 0.0001 to 0.00001 over the course of training. In both stages, training is performed using Stochastic Gradient Descent (SGD) as the optimization algorithm.







\begin{figure}[htbp]
\centering

\includegraphics[width=0.9\linewidth]{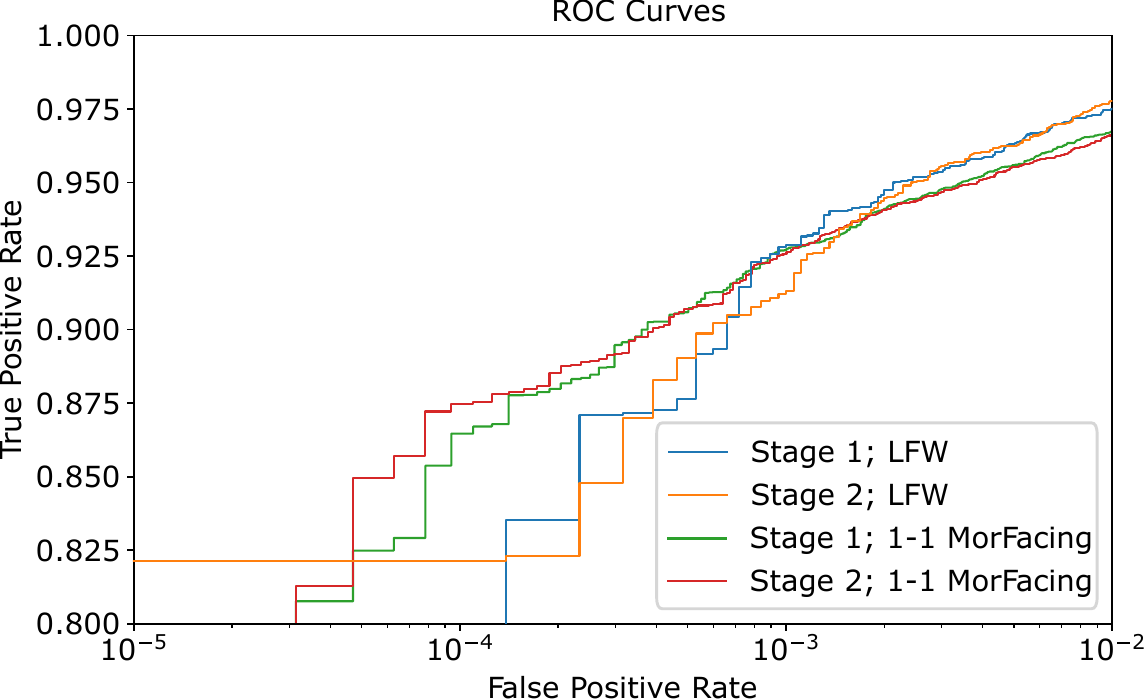}

\caption{ROC curves for the 1-1 verification performance on LFW and 1-1 MorFacing protocol}
\label{fig:MG_11_protocols_2_stages}
\end{figure}



We compare the performance of Stage 1 and Stage 2 models using several evaluation metrics and performance curves. As shown in Figure \ref{fig:MG_11_protocols_2_stages} and Table \ref{tab:MG_11_table_results_adaptation}, face recognition performance on in-the-wild images remained largely unchanged in Stage 2. However, a noticeable improvement was observed on the 1-1 MorFacing protocol, which consists of images of better quality. Indeed this may be explained by the quality filtering process of our training data.

\begin{figure}[htbp]
\centering

\includegraphics[width=0.9\linewidth]{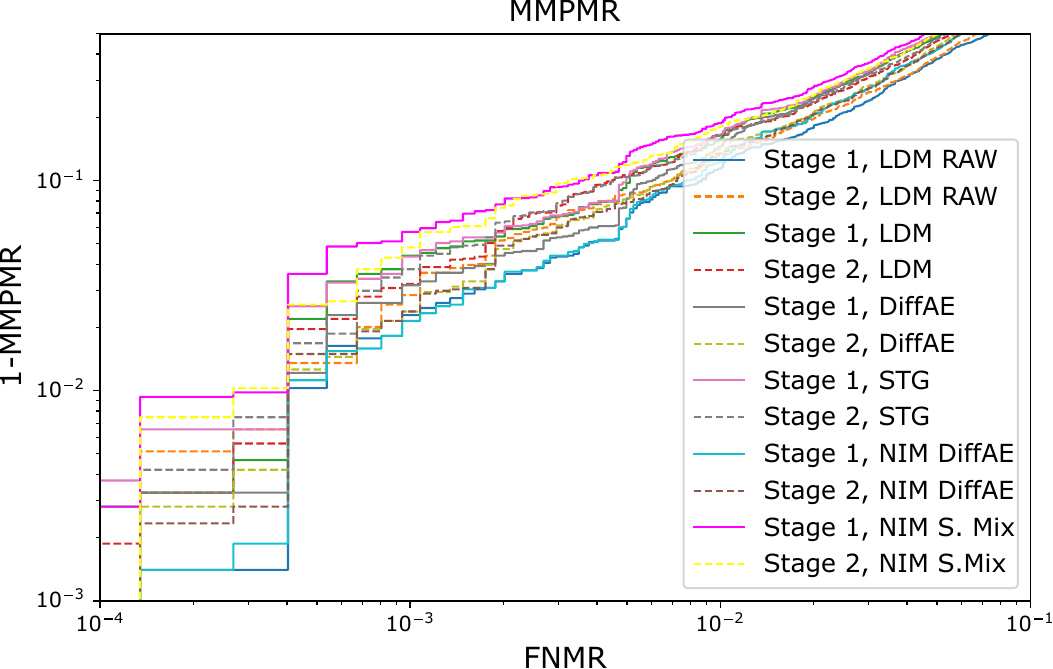}

\caption{MMPMR for face recognition model on 2 stages (Initial training ($m_{MG} = 0$ , $m_{BF} = 0.5$), Adaptation training ($m_{MG} = -0.1$ , $m_{BF} = 0.5$)) and protocols as a dependency from FNMR. Vertical axis is plotted as 1-MMPMR  (higher values \textrightarrow better robustness)}
\label{fig:MG_MMPMR_2stages}
\end{figure}

\begin{figure}[htbp]
\centering

\includegraphics[width=0.9\linewidth]{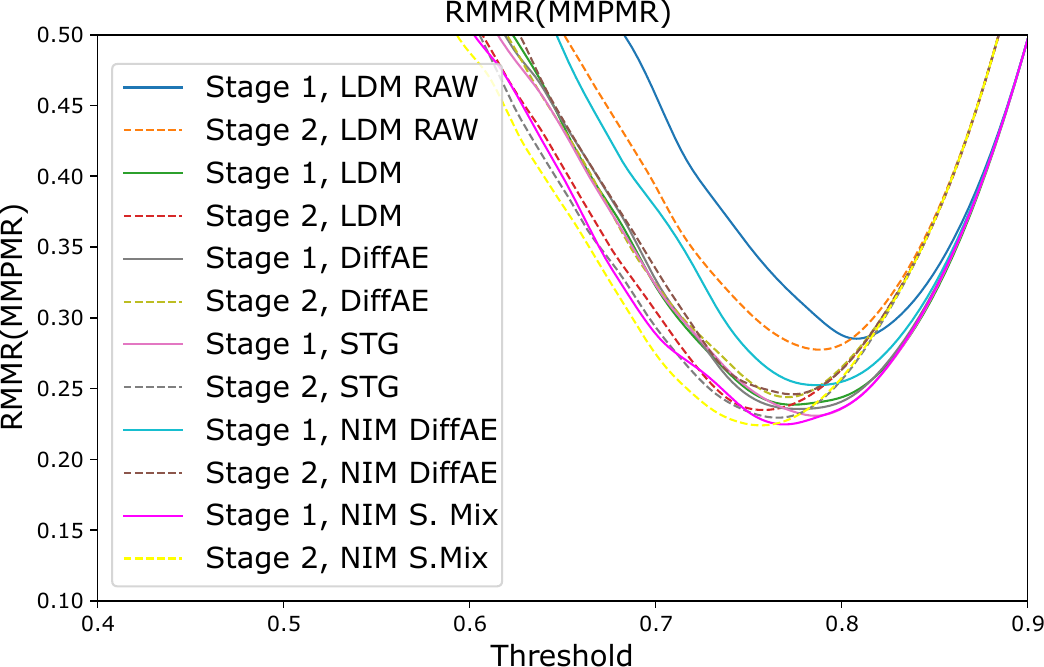}

\caption{RMMR for different models and protocols as a dependency from similarity threshold.}
\label{fig:MG_RMMR_2stages}
\end{figure}

Furthermore, the morphing robustness metrics ( MMPMR and RMMR) presented in Figures \ref{fig:MG_MMPMR_2stages} and \ref{fig:MG_RMMR_2stages}, and summarized in Tables \ref{tab:MG_RMMR_results_adaptation} and \ref{tab:MG_MMPMR_results_adaptation}, demonstrate a modest increase in robustness in most evaluation protocols. These findings validate the effectiveness of the proposed adaptation strategy and highlight its potential for enhancing already pretrained face recognition models.

\begin{table}[]
\caption{1-1 Verification performance for face recognition model at two stages of training: Initial training ($m_{MG} = 0$ , $m_{BF} = 0.5$), Adaptation training ($m_{MG} = -0.1$ , $m_{BF} = 0.5$). Performance is presented as  FNMR@FMR = [0.001, 0.0001] on two protocols - 1-1 MorFacing(MF) and LFW.}
\vspace{5px}
\resizebox{0.49\textwidth}{!}{
\begin{tabular}{|c|ll|ll|}
\hline
\multirow{2}{*}{\begin{tabular}[c]{@{}c@{}}\textit{1-1 Verification}\\ \textit{Performance}\end{tabular}} & \multicolumn{2}{c|}{\textit{1-1 MF; FMR =}}                & \multicolumn{2}{c|}{\textit{LFW; FMR =}}     \\ \cline{2-5} 
                                                                                           & \multicolumn{1}{c|}{\textit{0.001}} & \multicolumn{1}{c|}{\textit{0.0001}} & \multicolumn{1}{c|}{\textit{0.001}} & \textit{0.0001} \\ \hline
Stage 1 (initial training)                                                                                       & \multicolumn{1}{l|}{0.277}      &       0.642                   & \multicolumn{1}{l|}{0.128}      &    0.207   \\ \hline
Stage 2 (adaptation training)                                                                                      & \multicolumn{1}{l|}{0.305}      &          0.665                  & \multicolumn{1}{l|}{0.13}      &   0.178     \\ \hline
\end{tabular}}
\label{tab:MG_11_table_results_adaptation}
\end{table}

\begin{table}[]
\caption{The mimimum values of RMMR for different model on 2 stages of training (Stage 1. Initial training ($m_{MG} = 0$ , $m_{BF} = 0.5$), Stage 2. Adaptation training ($m_{MG} = -0.1$ , $m_{BF} = 0.5$)) in different testing protocols.}
\vspace{5px}
\centering
\begin{tabular}{|cl|cc|}
\hline
\multicolumn{2}{|c|}{\multirow{2}{*}{\begin{tabular}[c]{@{}c@{}}\textit{Minimum of } \\ \textit{RMMR curves}\end{tabular}}} & \multicolumn{2}{c|}{Model  }                                                                                                                                        \\ \cline{3-4} 
\multicolumn{2}{|c|}{}                                                                                        & \multicolumn{1}{|c|}{\textit{Stage 1}} & \textit{Stage 2} \\ \hline
\multicolumn{1}{|c|}{\multirow{6}{*}{\rotatebox[origin=c]{90}{Protocols}}}                           & LDM Raw                                  & \multicolumn{1}{l|}{0.285}    &  0.277   \\ \cline{2-4} 
\multicolumn{1}{|c|}{}                                                 & LDM                                  & \multicolumn{1}{l|}{0.238}    & 0.234    \\ \cline{2-4} 
\multicolumn{1}{|c|}{}                                                 & DiffAE                               & \multicolumn{1}{l|}{0.235}    & 0.243    \\ \cline{2-4} 
\multicolumn{1}{|c|}{}                                                 & STG                                  & \multicolumn{1}{l|}{0.230}    &  0.229   \\ \cline{2-4} 
\multicolumn{1}{|c|}{}                                                 & NIM S.Mix                            & \multicolumn{1}{l|}{0.225}    & 0.224    \\ \cline{2-4} 
\multicolumn{1}{|c|}{}                                                 & NIM DiffAE                           & \multicolumn{1}{l|}{0.252}    & 0.245   \\ \hline
\end{tabular}
\label{tab:MG_RMMR_results_adaptation}
\end{table}

\begin{table}[]
\caption{MMPMR values for the models on 2 stages of training (Initial training ($m_{MG} = 0$ , $m_{BF} = 0.5$), Adaptation training ($m_{MG} = -0.1$ , $m_{BF} = 0.5$)) in different testing protocols.}
\vspace{5px}
\centering
\resizebox{0.47\textwidth}{!}{
\begin{tabular}{|c|l|c|c|c|c|}
\hline
\multicolumn{2}{|c|}{\multirow{2}{*}{\begin{tabular}[c]{@{}c@{}}MMPMR \\ \textit{vs} FNMR\end{tabular}}} & \multicolumn{2}{c|}{FNMR = 0.01} & \multicolumn{2}{c|}{FNMR = 0.001} \\ \cline{3-6} 
\multicolumn{2}{|c|}{} & \textit{Stage 1} & \textit{Stage 2} & \textit{Stage 1} & \textit{Stage 2} \\ \hline
\multirow{6}{*}{\rotatebox[origin=c]{90}{Protocols}} & LDM Raw         & 0.886  & 0.868 & 0.981 & 0.974  \\ \cline{2-6} 
                                                     & LDM             & 0.844  & 0.835 & 0.962 & 0.969  \\ \cline{2-6} 
                                                     & DiffAE          & 0.855  & 0.859 & 0.973 & 0.978 \\ \cline{2-6} 
                                                     & STG             & 0.835  & 0.836 & 0.964 & 0.965  \\ \cline{2-6} 
                                                     & NIM S.Mix       & 0.812  & 0.820 & 0.948 & 0.957 \\ \cline{2-6} 
                                                     & NIM DiffAE      & 0.878  & 0.866 & 0.981 & 0.978  \\ \hline
\end{tabular}}
\label{tab:MG_MMPMR_results_adaptation}
\end{table}

\subsection{Feature Distribution Experiment}







In this subsection, we intend to analyze and compare the feature space distribution of morph samples across different trained models. Specifically, we investigate how morphs are positioned in the feature domain by generating visualizations similar to Fig. \ref{fig:MG_feature_distribution_morphguard}, but based on real high-dimensional features extracted from models.

To accomplish this, we use the MMPMR protocol of the MorFacing benchmark. However the elements of this protocol does not provide sufficient amount of morphs generated for a single identity subjects pair. That is why we approach our task by feature transformation techniques. For each element of the MMPMR protocol we construct a triplet, which consists of two original bona fide samples and a corresponding morph. Next, we perform a transformation of each triplet sample features.

For visualization, the 512-dimensional feature vectors of samples are projected into 2D by averaging values at even and odd indices. To align the features for comparison, we compute a rigid transformation based on the pair of original samples in each triplet and apply this transformation to all three features in the triplet. Namely this is a transformation between [$(X_1,Y_1)$;$(X_2,Y_2)$] and [$(-0.5, -0.5)$; $(0.5, 0.5)$], where $X_i$ $Y_i$ are the 2D features of the original samples.  This step ensures that the original samples align along a common axis, preserving relative distances while avoiding scaling.

This procedure allows us to build the spatial distribution of morph features in relation to their source identities. To quantify this distribution across different trained models, we compute the confidence ellipse at the 0.9 level for the morph samples, capturing the compactness and orientation of their spread in the feature space. 
Table \ref{tab:MG_feature_distr_ellipses} presents the values of the ellipse size, defined as $S = (W+H)/2$ where $W$ and $H$ denote the width and height of the ellipse. This size serves as an indicator of the spread of the morph sample distributions.

\begin{figure}[htbp]
\centering

\includegraphics[width=0.7\linewidth]{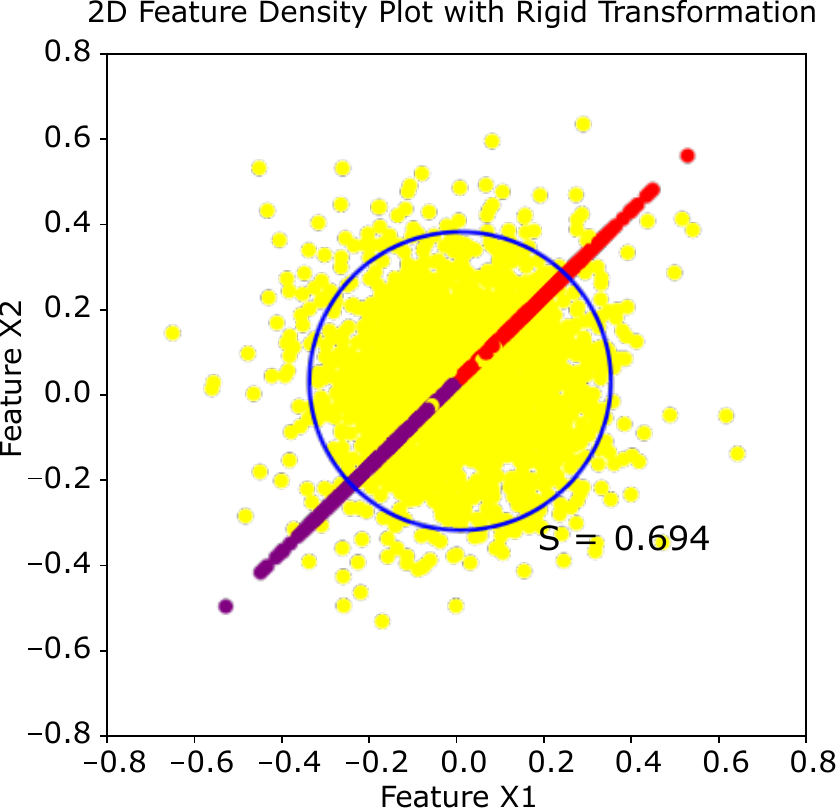}

a

\vspace{4px}
\includegraphics[width=0.7\linewidth]{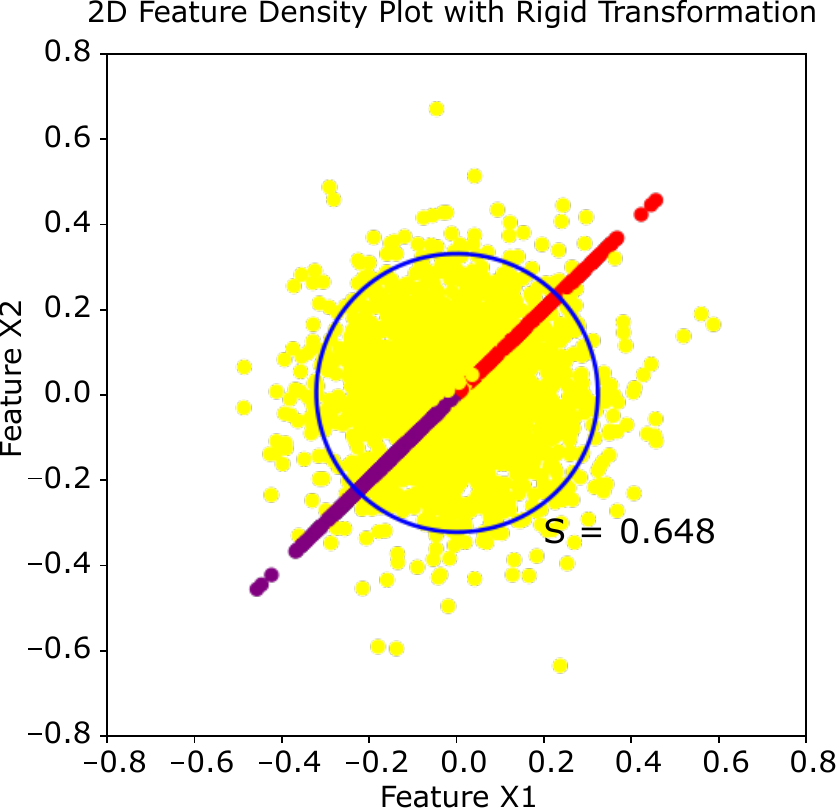}

b

\caption{Example transformed feature distributions for two models ( a - $m_{MG} = 0.0$ , $m_{BF} = 0.5$;  b - $m_{MG} = -0.1$ , $m_{BF} = 0.5$) on samples in protocol DiffAE. Model with $m_{MG} = -0.1$ lead to the more compact feature distribution of morph samples in comparison to a model with $m_{MG} = 0.0$}
\label{fig:MG_feature distributions_diffae}
\end{figure}

\begin{table}[]
\caption{MMPMR values for the models trained for estimation of the margin balance.}
\vspace{5px}
\resizebox{0.49\textwidth}{!}{\begin{tabular}{|l|ccccccccc|}
\hline
\multirow{2}{*}{\begin{tabular}[c]{@{}l@{}}Feature Distribution \\ vs. Protocols\end{tabular}} & \multicolumn{9}{c|}{Ellipse Size $S$}                                                                                                                                                                                                       \\ \cline{2-10} 
                                                                                                 & \multicolumn{1}{c|}{0.1} & \multicolumn{1}{c|}{0.05} & \multicolumn{1}{c|}{0.0} & \multicolumn{1}{c|}{-0.05} & \multicolumn{1}{c|}{-0.1} & \multicolumn{1}{c|}{-0.2} & \multicolumn{1}{c|}{-0.3} & \multicolumn{1}{c|}{Stg. 1} & Stg. 2 \\ \hline
LDM Raw                                                                                          & \multicolumn{1}{c|}{0.615}    & \multicolumn{1}{c|}{0.517}     & \multicolumn{1}{c|}{0.563}    & \multicolumn{1}{c|}{0.674}      & \multicolumn{1}{c|}{0.558}     & \multicolumn{1}{c|}{0.497}     & \multicolumn{1}{c|}{0.741}     & \multicolumn{1}{c|}{0.554}       &     0.548   \\ \hline
LDM                                                                                              & \multicolumn{1}{c|}{0.647}    & \multicolumn{1}{c|}{0.583}     & \multicolumn{1}{c|}{0.591}    & \multicolumn{1}{c|}{0.710}      & \multicolumn{1}{c|}{0.594}     & \multicolumn{1}{c|}{0.553}     & \multicolumn{1}{c|}{0.734}     & \multicolumn{1}{c|}{0.610}       &    0.590    \\ \hline
DiffAE                                                                                           & \multicolumn{1}{c|}{0.775}    & \multicolumn{1}{c|}{0.628}     & \multicolumn{1}{c|}{0.694}    & \multicolumn{1}{c|}{0.749}      & \multicolumn{1}{c|}{0.648}     & \multicolumn{1}{c|}{0.615}     & \multicolumn{1}{c|}{0.779}     & \multicolumn{1}{c|}{0.612}       &    0.609    \\ \hline
STG                                                                                              & \multicolumn{1}{c|}{0.711}    & \multicolumn{1}{c|}{0.641}     & \multicolumn{1}{c|}{0.668}    & \multicolumn{1}{c|}{0.751}      & \multicolumn{1}{c|}{0.651}     & \multicolumn{1}{c|}{0.612}     & \multicolumn{1}{c|}{0.791}     & \multicolumn{1}{c|}{0.724}       &    0.679    \\ \hline
NIM. S. Mix                                                                                      & \multicolumn{1}{c|}{0.710}    & \multicolumn{1}{c|}{0.625}     & \multicolumn{1}{c|}{0.652}    & \multicolumn{1}{c|}{0.692}      & \multicolumn{1}{c|}{0.669}     & \multicolumn{1}{c|}{0.576}     & \multicolumn{1}{c|}{0.698}     & \multicolumn{1}{c|}{0.661}       &    0.626    \\ \hline
NIM. DiffAE                                                                                      & \multicolumn{1}{c|}{0.660}    & \multicolumn{1}{c|}{0.590}     & \multicolumn{1}{c|}{0.707}    & \multicolumn{1}{c|}{0.727}      & \multicolumn{1}{c|}{0.583}     & \multicolumn{1}{c|}{0.607}     & \multicolumn{1}{c|}{0.792}     & \multicolumn{1}{c|}{0.587}       &    0.643    \\ \hline
\end{tabular}}
\label{tab:MG_feature_distr_ellipses}
\end{table}


The results presented in Table \ref{tab:MG_feature_distr_ellipses} correspond to the models evaluated in Subsections \ref{section:margin_balance} and \ref{section:model_adaptation}. They show that, for most models with a non-zero margin parameter $m_{MG}$, the distribution of morph samples is more compact compared to the baseline model with $m_{MG} = 0.0$. Notably, models with a large absolute margin tend to produce more dispersed morph distributions. Furthermore, a comparison between Stage 1 and Stage 2 models from Subsection \ref{section:model_adaptation} indicates that adapting a pretrained model results in a denser distribution of morph samples.

\section{Conclusion}

In this work, we propose a novel approach for training deep networks for face recognition with increased robustness to face morphing attacks. Our approach is based on modifying the classification task and adapting it to allow training with images of face morphs. We propose a dual-branch classification strategy designed to address the ambiguity in the labeling of face morphs.  

Our strategy has demonstrated its effectiveness on public benchmarks for evaluating robustness to face morphing attacks. We conducted extensive experiments to find the optimal balance for the components of our loss function, which results in the best robustness.

Moreover, our approach is universal and can be used to modify existing strategies for face recognition via classification, which was proved with experimental results.

We also demonstrated that the application of our approach gives a more compact distribution of morph samples.

In our future work, we will extend these results for different marginal softmax loss functions and will consider a sample specific modifications of our strategy.


{\small
\bibliographystyle{IEEEtran}
\bibliography{egbib}

\begin{thebibliography}{10}
\providecommand{\url}[1]{#1}
\csname url@samestyle\endcsname
\providecommand{\newblock}{\relax}
\providecommand{\bibinfo}[2]{#2}
\providecommand{\BIBentrySTDinterwordspacing}{\spaceskip=0pt\relax}
\providecommand{\BIBentryALTinterwordstretchfactor}{4}
\providecommand{\BIBentryALTinterwordspacing}{\spaceskip=\fontdimen2\font plus
\BIBentryALTinterwordstretchfactor\fontdimen3\font minus \fontdimen4\font\relax}
\providecommand{\BIBforeignlanguage}[2]{{%
\expandafter\ifx\csname l@#1\endcsname\relax
\typeout{** WARNING: IEEEtran.bst: No hyphenation pattern has been}%
\typeout{** loaded for the language `#1'. Using the pattern for}%
\typeout{** the default language instead.}%
\else
\language=\csname l@#1\endcsname
\fi
#2}}
\providecommand{\BIBdecl}{\relax}
\BIBdecl

\bibitem{torkar2023morphing}
M.~Torkar, ``Morphing cases in slovenia,'' {NIST IFPS}, 2022, ministry of the Interior Police, Slovenia.

\bibitem{ImageNet_cite}
O.~Russakovsky, J.~Deng, H.~Su, J.~Krause, S.~Satheesh, S.~Ma, Z.~Huang, A.~Karpathy, A.~Khosla, M.~B., A.~C. Berg, and L.~Fei-Fei, ``{{ImageNet Large Scale Visual Recognition Challenge}},'' \emph{{IJCV}}, vol. 115, no.~3, pp. 211--252, 2015.

\bibitem{chopra_metric_paper}
S.~Chopra, R.~Hadsell, and Y.~LeCun, ``Learning a similarity metric discriminatively, with application to face verification,'' \emph{2005 IEEE Computer Society Conference on Computer Vision and Pattern Recognition (CVPR'05)}, vol.~1, pp. 539--546 vol. 1, 2005.

\bibitem{facenet}
F.~{Schroff}, D.~{Kalenichenko}, and J.~{Philbin}, ``{FaceNet: A unified embedding for face recognition and clustering},'' in \emph{2015 IEEE Conference CVPR}, 2015, pp. 815--823.

\bibitem{working_hard}
A.~Mishchuk, D.~Mishkin, F.~Radenovi\'{c}, and J.~Matas, ``Working hard to know your neighbor's margins: local descriptor learning loss,'' in \emph{Proceedings of the 31st International Conference on Neural Information Processing Systems}, ser. NIPS'17.\hskip 1em plus 0.5em minus 0.4em\relax Red Hook, NY, USA: Curran Associates Inc., 2017, p. 4829–4840.

\bibitem{coreface}
Y.~Song and F.~Wang, ``Coreface: Sample-guided contrastive regularization for deep face recognition,'' 2023.

\bibitem{deepid_paper}
Y.~{Sun}, X.~{Wang}, and X.~{Tang}, ``{Deep Learning Face Representation from Predicting 10,000 Classes},'' in \emph{2014 IEEE Conference on Computer Vision and Pattern Recognition}, 2014, pp. 1891--1898.

\bibitem{deepid2_paper}
Y.~Sun, Y.~Chen, X.~Wang, and X.~Tang, ``{Deep Learning Face Representation by Joint Identification-Verification},'' in \emph{NIPS}, 2014.

\bibitem{deepid2_plus_paper}
Y.~Sun, X.~Wang, and X.~Tang, ``Deeply learned face representations are sparse, selective, and robust,'' \emph{2015 IEEE Conference on CVPR}, pp. 2892--2900, 2015.

\bibitem{centerface_paper}
Y.~Wen, K.~Zhang, Z.~Li, and Y.~Qiao, ``{A Discriminative Feature Learning Approach for Deep Face Recognition},'' in \emph{Computer Vision -- ECCV 2016}, 2016, pp. 499--515.

\bibitem{sphereface_paper}
W.~{Liu}, Y.~{Wen}, Z.~{Yu}, M.~{Li}, B.~{Raj}, and L.~{Song}, ``{SphereFace: Deep Hypersphere Embedding for Face Recognition},'' in \emph{2017 IEEE Conference on Computer Vision and Pattern Recognition (CVPR)}, 2017, pp. 6738--6746.

\bibitem{cosface_paper}
H.~{Wang}, Y.~{Wang}, Z.~{Zhou}, X.~{Ji}, D.~{Gong}, J.~{Zhou}, Z.~{Li}, and W.~{Liu}, ``{CosFace: Large Margin Cosine Loss for Deep Face Recognition},'' in \emph{2018 IEEE/CVF Conference on Computer Vision and Pattern Recognition}, 2018, pp. 5265--5274.

\bibitem{arcface_paper}
J.~{Deng}, J.~{Guo}, N.~{Xue}, and S.~{Zafeiriou}, ``{ArcFace: Additive Angular Margin Loss for Deep Face Recognition},'' in \emph{2019 Conference on CVPR}, 2019, pp. 4685--4694.

\bibitem{npcface}
\BIBentryALTinterwordspacing
D.~Zeng, H.~Shi, H.~Du, J.~Wang, Z.~Lei, and T.~Mei, ``{NPCFace: {A} Negative-Positive Cooperation Supervision for Training Large-scale Face Recognition},'' \emph{CoRR}, vol. abs/2007.10172, 2020. [Online]. Available: \url{https://arxiv.org/abs/2007.10172}
\BIBentrySTDinterwordspacing

\bibitem{probabilistic_embedding}
Y.~{Shi} and A.~{Jain}, ``{Probabilistic Face Embeddings},'' in \emph{{2019 IEEE/CVF International Conference on Computer Vision (ICCV)}}, 2019, pp. 6901--6910.

\bibitem{Magface}
Q.~Meng, S.~Zhao, Z.~Huang, and F.~Zhou, ``Magface: A universal representation for face recognition and quality assessment,'' in \emph{2021 IEEE/CVF Conference on CVPR}, 2021, pp. 14\,220--14\,229.

\bibitem{QualFace}
J.~Tremoço, I.~Medvedev, and N.~Gonçalves, ``{QualFace: Adapting Deep Learning Face Recognition for ID and Travel Documents with Quality Assessment},'' in \emph{2021 International Conference of the BIOSIG}, 2021, pp. 1--6.

\bibitem{QualFace2}
I.~Medvedev, J.~Tremoço, B.~Mano, L.~E. Santo, and N.~Gonçalves, ``Towards understanding the character of quality sampling in deep learning face recognition,'' \emph{IET Biometrics}, vol.~11, no.~5, pp. 498--511, 2022.

\bibitem{AdaFace}
M.~Kim, A.~K. Jain, and X.~Liu, ``Adaface: Quality adaptive margin for face recognition,'' \emph{2022 IEEE/CVF Conference on CVPR}, pp. 18\,729--18\,738, 2022.

\bibitem{magic_passport}
M.~Ferrara, A.~Franco, and D.~Maltoni, ``{The magic passport},'' \emph{{IJCB 2014 - 2014 IEEE/IAPR}}, 12 2014.

\bibitem{ubo_morpher}
{Biometric System Laboratory}, ``{UBO-Morpher},'' 2018, http://biolab.csr.unibo.it/Research.asp. (accessed: September 1, 2022).

\bibitem{NIPS2014_5ca3e9b1}
I.~J. Goodfellow, J.~P.Abadie, M.~Mirza, B.~Xu, D.~Warde-Farley, S.~Ozair, A.~Courville, and Y.~Bengio, ``Generative adversarial nets,'' in \emph{NeurIPS}.\hskip 1em plus 0.5em minus 0.4em\relax Curran Associates, Inc., 2014.

\bibitem{morGAN}
N.~{Damer}, A.~M. {Saladié}, A.~{Braun}, and A.~{Kuijper}, ``{MorGAN: Recognition Vulnerability and Attack Detectability of Face Morphing Attacks Created by Generative Adversarial Network},'' in \emph{2018 IEEE 9th International Conference on BTAS}, 2018, pp. 1--10.

\bibitem{styleGAN}
T.~{Karras}, S.~{Laine}, and T.~{Aila}, ``{A Style-Based Generator Architecture for Generative Adversarial Networks},'' in \emph{2019 IEEE/CVF Conference on CVPR}, 2019, pp. 4396--4405.

\bibitem{MIPGAN_morphing_paper}
\BIBentryALTinterwordspacing
H.~Zhang, S.~K. Venkatesh, R.~Ramachandra, K.~B. Raja, N.~Damer, and C.~Busch, ``Mipgan—generating strong and high quality morphing attacks using identity prior driven gan,'' \emph{IEEE Transactions on Biometrics, Behavior, and Identity Science}, vol.~3, pp. 365--383, 2021. [Online]. Available: \url{https://api.semanticscholar.org/CorpusID:234913212}
\BIBentrySTDinterwordspacing

\bibitem{morcode}
A.~Reddy, R.~Ramachandra, K.~Sreenivasa, and P.~Mitra, ``Morcode: Face morphing attack generation using generative codebooks,'' in \emph{In Proceedings of 35th British Machine Vision Conference (BMVC 2024)}, 2024.

\bibitem{MorDIFF}
N.~Damer, M.~Fang, P.~Siebke, J.~N. Kolf, M.~Huber, and F.~Boutros, ``{MorDIFF: Recognition Vulnerability and Attack Detectability of Face Morphing Attacks Created by Diffusion Autoencoders},'' in \emph{In Proceedings of 11th International Workshop on Biometrics and Forensics (IWBF)}, 2023.

\bibitem{greedydim}
Z.~W. Blasingame and C.~Liu, ``Greedy-dim: Greedy algorithms for unreasonably effective face morphs,'' in \emph{2024 IEEE International Joint Conference on Biometrics (IJCB)}, 2024, pp. 1--11.

\bibitem{ladimo}
M.~Grimmer and C.~Busch, ``Ladimo: Face morph generation through biometric template inversion with latent diffusion,'' in \emph{2024 IEEE International Joint Conference on Biometrics (IJCB)}, 2024, pp. 1--7.

\bibitem{ReGenMorph}
N.~Damer, K.~B. Raja, M.~Sussmilch, S.~K. Venkatesh, F.~Boutros, M.~Fang, F.~Kirchbuchner, R.~Ramachandra, and A.~Kuijper, ``Regenmorph: Visibly realistic gan generated face morphing attacks by attack re-generation,'' in \emph{ISVC}, 2021.

\bibitem{7791169}
R.~Raghavendra, K.~B. Raja, and C.~Busch, ``Detecting morphed face images,'' in \emph{2016 IEEE 8th International Conference on BTAS}, Sep. 2016, pp. 1--7.

\bibitem{OJALA199651}
T.~Ojala, M.~Pietik{\"a}inen, and D.~Harwood, ``A comparative study of texture measures with classification based on featured distributions,'' \emph{Pattern recognition}, vol.~29, no.~1, pp. 51--59, 1996.

\bibitem{Blur_Texture_Classification}
V.~Ojansivu and J.~Heikkil\"{a}, ``Blur insensitive texture classification using local phase quantization,'' in \emph{Springer-Verlag}, ser. ICISP '08, Berlin, Heidelberg, 2008, p. 236–243.

\bibitem{Morphing_detection_PRNU}
U.~{Scherhag}, L.~{Debiasi}, C.~{Rathgeb}, C.~{Busch}, and A.~{Uhl}, ``{Detection of Face Morphing Attacks Based on PRNU Analysis},'' \emph{{IEEE T-BIOM}}, vol.~1, no.~4, pp. 302--317, 2019.

\bibitem{orthomad}
P.~C. Neto, T.~Gonçalves, M.~Huber, N.~Damer, A.~F. Sequeira, and J.~S. Cardoso, ``Orthomad: Morphing attack detection through orthogonal identity disentanglement,'' in \emph{BIOSIG 2022}, pp. 1--5.

\bibitem{MorDeephy}
I.~Medvedev, F.~Shadmand, and N.~Gonçalves, ``Mordeephy: Face morphing detection via fused classification,'' in \emph{Proceedings of ICPRAM}.\hskip 1em plus 0.5em minus 0.4em\relax SciTePress, 2023, pp. 193--204.

\bibitem{madation}
\BIBentryALTinterwordspacing
E.~Caldeira, G.~Ozgur, T.~Chettaoui, M.~Ivanovska, P.~Peer, F.~Boutros, V.~Struc, and N.~Damer, ``{ MADation: Face Morphing Attack Detection with Foundation Models },'' in \emph{2025 IEEE/CVF Winter Conference on Applications of Computer Vision Workshops (WACVW)}.\hskip 1em plus 0.5em minus 0.4em\relax Los Alamitos, CA, USA: IEEE Computer Society, Mar. 2025, pp. 1565--1575. [Online]. Available: \url{https://doi.ieeecomputersociety.org/10.1109/WACVW65960.2025.00179}
\BIBentrySTDinterwordspacing

\bibitem{clip}
\BIBentryALTinterwordspacing
A.~Radford, J.~W. Kim, C.~Hallacy, A.~Ramesh, G.~Goh, S.~Agarwal, G.~Sastry, A.~Askell, P.~Mishkin, J.~Clark, G.~Krueger, and I.~Sutskever, ``Learning transferable visual models from natural language supervision,'' in \emph{Proceedings of the 38th International Conference on Machine Learning, {ICML} 2021, 18-24 July 2021, Virtual Event}, ser. Proceedings of Machine Learning Research, M.~Meila and T.~Zhang, Eds., vol. 139.\hskip 1em plus 0.5em minus 0.4em\relax {PMLR}, 2021, pp. 8748--8763. [Online]. Available: \url{http://proceedings.mlr.press/v139/radford21a.html}
\BIBentrySTDinterwordspacing

\bibitem{Tapia2021SingleMA}
J.~E. Tapia and C.~Busch, ``Single morphing attack detection using feature selection and visualization based on mutual information,'' \emph{IEEE Access}, vol.~9, pp. 167\,628--167\,641, 2021.

\bibitem{Double_Siamese_Morphing}
G.~Borghi, E.~Pancisi, M.~Ferrara, and D.~Maltoni, ``A double siamese framework for differential morphing attack detection,'' \emph{Sensors}, vol.~21, p. 3466, 05 2021.

\bibitem{FMD_Feature_Wise_Supervision}
L.~Qin, F.~Peng, and M.~Long, ``Face morphing attack detection and localization based on feature-wise supervision,'' \emph{IEEE TIFS}, vol.~17, pp. 3649--3662, 2022.

\bibitem{dmad_combining_identity_features}
N.~D. Domenico, G.~Borghi, A.~Franco, and D.~Maltoni, ``Combining identity features and artifact analysis for differential morphing attack detection,'' in \emph{ICIAP (1)}, 2023, pp. 100--111.

\bibitem{face_demorphing}
M.~{Ferrara}, A.~{Franco}, and D.~{Maltoni}, ``Face demorphing,'' \emph{IEEE Transactions on Information Forensics and Security}, vol.~13, no.~4, pp. 1008--1017, 2018.

\bibitem{iftnet}
\BIBentryALTinterwordspacing
L.-B. Zhang, S.~Chen, M.~Long, and J.~Cai, ``Face de-morphing based on identity feature transfer,'' \emph{IET Image Processing}, vol.~19, no.~1, p. e13324, 2025. [Online]. Available: \url{https://ietresearch.onlinelibrary.wiley.com/doi/abs/10.1049/ipr2.13324}
\BIBentrySTDinterwordspacing

\bibitem{Biometric_Systems_under_Morphing_Attacks}
U.~Scherhag, A.~Nautsch, C.~Rathgeb, M.~Gomez-Barrero, R.~Veldhuis, L.~Spreeuwers, M.~Schils, D.~Maltoni, P.~Grother, S.~Marcel, R.~Breithaupt, R.~Ramachandra, and C.~Busch, ``\BIBforeignlanguage{English}{Biometric systems under morphing attacks: Assessment of morphing techniques and vulnerability reporting},'' in \emph{\BIBforeignlanguage{English}{2017 International Conference of the Biometrics Special Interest Group, BIOSIG 2017}}, 9 2017.

\bibitem{Marriott2020RobustnessOF}
\BIBentryALTinterwordspacing
R.~T. Marriott, S.~Romdhani, S.~Gentric, and L.~Chen, ``Robustness of facial recognition to gan-based face-morphing attacks,'' \emph{ArXiv}, vol. abs/2012.10548, 2020. [Online]. Available: \url{https://api.semanticscholar.org/CorpusID:229339565}
\BIBentrySTDinterwordspacing

\bibitem{medvedev2024quadruplet}
I.~Medvedev and N.~Gonçalves, ``Quadruplet loss for improving the robustness to face morphing attacks,'' in \emph{2024 12th International Workshop on Biometrics and Forensics (IWBF)}, 2024, pp. 01--06.

\bibitem{bench_NIST_morph}
{NIST}, ``{FRVT MORPH},'' 2022, https://pages.nist.gov/frvt/html/frvt\_morph.html.

\bibitem{FRGC_V2_dataset}
P.~J. Phillips, P.~Flynn, T.~Scruggs, K.~Bowyer, J.~K. Chang, K.~Hoffman, J.~Marques, J.~Min, and W.~Worek, ``{Overview of the Face Recognition Grand Challenge},'' in \emph{{Proceedings of IEEE Computer Society Conference on Computer Vision and Pattern Recognition}}, vol.~1, 07 2005, pp. 947-- 954.

\bibitem{iwbf1}
I.~Batskos and L.~Spreeuwers, ``Improving fully automated landmark-based face morphing,'' in \emph{In Proceedings of the 12th International Workshop On Biometrics And Forensics (IWBF), 2024}, 2024.

\bibitem{iwbf2}
N.~D. Domenico, G.~Borghi, A.~Franco, and D.~Maltoni, ``Face restoration for morphed images,'' in \emph{In Proceedings of the 12th International Workshop On Biometrics And Forensics (IWBF), 2024}, 2024.

\bibitem{morfacing}
I.~Medvedev and N.~Gonçalves, ``Morfacing: A benchmark for estimation face recognition robustness to face morphing attacks,'' in \emph{2024 IEEE International Joint Conference on Biometrics (IJCB)}, 2024, pp. 1--10.

\bibitem{WebFace260M}
Z.~Zhu, G.~Huang, J.~Deng, Y.~Ye, J.~Huang, X.~Chen, J.~Zhu, T.~Yang, J.~Lu, D.~Du, and J.~Zhou, ``{WebFace260M: A Benchmark Unveiling the Power of Million-Scale Deep Face Recognition},'' in \emph{Proceedings of the IEEE/CVF Conference on Computer Vision and Pattern Recognition (CVPR)}, June 2021, pp. 10\,492--10\,502.

\bibitem{quality_driven}
I.~Medvedev and N.~Gonçalves, ``Improving performance of facial biometrics with quality-driven dataset filtering,'' in \emph{2023 IEEE 17th International Conference on Automatic Face and Gesture Recognition (FG)}, 2023, pp. 1--8.

\bibitem{variance_of_laplacian}
R.~Bansal, G.~Raj, and T.~Choudhury, ``Blur image detection using laplacian operator and open-cv,'' in \emph{2016 International Conference System Modeling Advancement in Research Trends (SMART)}, 2016, pp. 63--67.

\bibitem{faceQnet}
J.~{Hernandez-Ortega}, J.~{Galbally}, J.~{Fierrez}, R.~{Haraksim}, and L.~{Beslay}, ``{FaceQnet: Quality Assessment for Face Recognition based on Deep Learning},'' in \emph{2019 International Conference on Biometrics (ICB)}, 2019, pp. 1--8.

\bibitem{brisque}
A.~{Mittal}, A.~K. {Moorthy}, and A.~C. {Bovik}, ``No-reference image quality assessment in the spatial domain,'' \emph{IEEE Transactions on Image Processing}, vol.~21, no.~12, pp. 4695--4708, 2012.

\bibitem{fiiqa}
L.~Zhang, L.~Zhang, and L.~Li, ``{Illumination Quality Assessment for Face Images: A Benchmark and a Convolutional Neural Networks Based Model},'' \emph{Lecture Notes in Computer Science}, vol. 10636 LNCS, pp. 583--593, 2017.

\bibitem{pose}
N.~Ruiz, E.~Chong, and J.~Rehg, ``Fine-grained head pose estimation without keypoints,'' in \emph{The IEEE Conference on Conference on Computer Vision and Pattern Recognition (CVPR) Workshops}, 06 2018.

\bibitem{VGGface2}
Q.~Cao, L.~Shen, W.~Xie, O.~M. Parkhi, and A.~Zisserman, ``Vggface2: A dataset for recognising faces across pose and age,'' in \emph{International Conference on FG}, 2018.

\bibitem{casia_webface}
D.~Yi, Z.~Lei, S.~Liao, and S.~Li, ``Learning face representation from scratch,'' \emph{ArXiv}, vol. abs/1411.7923, 2014.

\bibitem{ms_celeb_face}
Y.~Guo, L.~Zhang, Y.~Hu, X.~He, and J.~Gao, ``{MS-Celeb-1M: A Dataset and Benchmark for Large-Scale Face Recognition},'' in \emph{Proceedings of ECCV}, vol. 9907, 10 2016, pp. 87--102.

\bibitem{MS-Celeb-1M_CleanList}
C.~Jin, R.~Jin, K.~Chen, and Y.~Dou, ``{A community detection approach to cleaning extremely large face database},'' \emph{{Computational intelligence and neuroscience}}, vol. 2018, 2018.

\bibitem{Partial_fc}
X.~An, X.~Zhu, Y.~Gao, Y.~Xiao, Y.~Zhao, Z.~Feng, L.~Wu, B.~Qin, M.~Zhang, D.~Zhang, and Y.~Fu, ``{Partial FC: Training 10 Million Identities on a Single Machine},'' in \emph{Proceedings of the IEEE/CVF International Conference on Computer Vision (ICCV) Workshops}, October 2021, pp. 1445--1449.

\bibitem{morphing_artifacts_retouching}
G.~Borghi, A.~Franco, G.~Graffieti, and D.~Maltoni, ``Automated artifact retouching in morphed images with attention maps,'' \emph{IEEE Access}, vol.~9, pp. 136\,561--136\,579, 2021.

\bibitem{diffae}
K.~Preechakul, N.~Chatthee, S.~Wizadwongsa, and S.~Suwajanakorn, ``Diffusion autoencoders: Toward a meaningful and decodable representation,'' in \emph{IEEE Conference on Computer Vision and Pattern Recognition (CVPR)}, 2022.

\bibitem{GuerraMG23}
\BIBentryALTinterwordspacing
C.~Guerra, J.~Marcos, and N.~Gon{\c{c}}alves, ``Automatic validation of {ICAO} compliance regarding head coverings: An inclusive approach concerning religious circumstances,'' in \emph{{BIOSIG} 2023,}.\hskip 1em plus 0.5em minus 0.4em\relax {IEEE}, 2023, pp. 1--4. [Online]. Available: \url{https://doi.org/10.1109/BIOSIG58226.2023.10345995}
\BIBentrySTDinterwordspacing

\bibitem{learn_opencv_morpher}
M.~Satya, ``{Face Morph Using OpenCV},'' 2016, www.learnopencv.com/face-morph-using-opencv-cpp-python/. (accessed: September 1, 2022).

\bibitem{Schardong_2024_CVPR}
G.~Schardong, T.~Novello, H.~Paz, I.~Medvedev, V.~da~Silva, L.~Velho, and N.~Gon\c{c}alves, ``Neural implicit morphing of face images,'' in \emph{Proceedings of the IEEE/CVF Conference on Computer Vision and Pattern Recognition (CVPR)}, June 2024, pp. 7321--7330.

\bibitem{LFW_dataset}
G.~B. Huang, M.~Ramesh, T.~Berg, and E.~Learned-Miller, ``{Labeled Faces in the Wild: A Database for Studying Face Recognition in Unconstrained Environments},'' Tech. Rep. 07-49, October 2007.

\end{thebibliography}

}

\end{document}